\documentclass[sigconf]{acmart}

\usepackage{latexsym}
\usepackage{amsmath}
\usepackage{graphicx}
\usepackage{subfigure}
\usepackage{float}
\usepackage{booktabs}
\usepackage{multirow}
\usepackage{color}
\usepackage{caption}
\usepackage{subcaption}
\usepackage{pifont}
\usepackage{balance}
\usepackage{hyperref}
\AtBeginDocument{%
  \providecommand\BibTeX{{%
    \normalfont B\kern-0.5em{\scshape i\kern-0.25em b}\kern-0.8em\TeX}}}

\copyrightyear{2024} 
\acmYear{2024} 
\setcopyright{acmlicensed}
\acmConference[ICMR '24]{Proceedings of the 2024 International Conference on Multimedia Retrieval}{June 10--14, 2024}{Phuket, Thailand}
\acmBooktitle{Proceedings of the 2024 International Conference on Multimedia Retrieval (ICMR '24), June 10--14, 2024, Phuket, Thailand}
\acmDOI{10.1145/3652583.3658049}
\acmISBN{979-8-4007-0619-6/24/06}

\begin{document}

%%
%% The "title" command has an optional parameter,
%% allowing the author to define a "short title" to be used in page headers.
\title{Progressive Multi-modal Conditional Prompt Tuning}

\author{Xiaoyu Qiu}
\email{qiuxy@mail.ustc.edu.cn}
\orcid{0000-0002-0966-2006}
\affiliation{%
  \institution{University of Science and Technology of China}
  \streetaddress{100 Fuxing Rd}
  \city{Hefei}
  \state{Anhui}
  \country{China}}

\author{Hao Feng}
\email{haof@mail.ustc.edu.cn}
\orcid{0000-0001-8127-6639}
\affiliation{%
  \institution{University of Science and Technology of China}
  \streetaddress{100 Fuxing Rd}
  \city{Hefei}
  \state{Anhui}
  \country{China}}

\author{Yuechen Wang}
\email{wyc9725@mail.ustc.edu.cn}
\orcid{0000-0003-0965-584X}
\affiliation{%
  \institution{University of Science and Technology of China}
  \streetaddress{100 Fuxing Rd}
  \city{Hefei}
  \state{Anhui}
  \country{China}}

\author{Wengang Zhou}
\email{zhwg@ustc.edu.cn}
\orcid{0000-0003-1690-9836}
\affiliation{%
  \institution{University of Science and Technology of China}
  \streetaddress{100 Fuxing Rd}
  \city{Hefei}
  \state{Anhui}
  \country{China}}

\author{Houqiang Li}
% \authornote{Corresponding authors: Houqiang Li.}
\email{lihq@ustc.edu.cn}
\orcid{0000-0003-2188-3028}
\affiliation{%
  \institution{University of Science and Technology of China}
  \streetaddress{100 Fuxing Rd}
  \city{Hefei}
  \state{Anhui}
  \country{China}}

%%
%% By default, the full list of authors will be used in the page
%% headers. Often, this list is too long, and will overlap
%% other information printed in the page headers. This command allows
%% the author to define a more concise list
%% of authors' names for this purpose.
\renewcommand{\shortauthors}{Xiaoyu Qiu, Hao Feng, Yuechen Wang, Wengang Zhou, \& Houqiang Li}

%%
%% The abstract is a short summary of the work to be presented in the
%% article.
\begin{abstract}
Pre-trained vision-language models (VLMs) have shown remarkable generalization capabilities via prompting, which leverages VLMs as knowledge bases to extract information beneficial for downstream tasks. However, existing methods primarily employ uni-modal prompting, which only engages a uni-modal branch, failing to simultaneously adjust vision-language (V-L) features. Additionally, the one-pass forward pipeline in VLM encoding struggles to align V-L features that have a huge gap. Confronting these challenges, we propose a novel method, Progressive Multi-modal conditional Prompt Tuning (ProMPT). ProMPT exploits a recurrent structure, optimizing and aligning V-L features by iteratively utilizing image and current encoding information. It comprises an initialization and a multi-modal iterative evolution (MIE) module. Initialization is responsible for encoding images and text using a VLM, followed by a feature filter that selects text features similar to image. MIE then facilitates multi-modal prompting through class-conditional vision prompting, instance-conditional text prompting, and feature filtering. In each MIE iteration, vision prompts are obtained from filtered text features via a vision generator, promoting image features to focus more on target object during vision prompting. The encoded image features are fed into a text generator to produce text prompts that are more robust to class shifts. Thus, V-L features are progressively aligned, enabling advance from coarse to exact prediction. Extensive experiments are conducted in three settings to evaluate the efficacy of ProMPT. The results indicate that ProMPT outperforms existing methods on average across all settings, demonstrating its superior generalization and robustness. Code is available at \href{https://github.com/qiuxiaoyu9954/ProMPT}{\textbf{https://github.com/qiuxiaoyu9954/ProMPT}}. 
\end{abstract}

%%
%% The code below is generated by the tool at http://dl.acm.org/ccs.cfm.
%% Please copy and paste the code instead of the example below.
%%


%%
%% Keywords. The author(s) should pick words that accurately describe
%% the work being presented. Separate the keywords with commas.

%% A "teaser" image appears between the author and affiliation
%% information and the body of the document, and typically spans the
%% page.

% \received{20 February 2024}
% \received[revised]{12 March 2024}
% \received[accepted]{5 June 2024}

%%
%% This command processes the author and affiliation and title
%% information and builds the first part of the formatted document.
\maketitle

\vspace{-0.1in}
\section{Introduction}
In recent years, the advent of pre-trained vision-language models (VLMs), such as CLIP \cite{radford2021learning} and ALIGN \cite{jia2021scaling}, has marked a significant leap in the field of computer vision (CV), particularly in terms of their generalizability in downstream tasks. VLMs are trained on large-scale aligned text-image pairs, enabling them to learn open-set visual concepts from natural language during pre-training. This approach significantly enhances their zero-shot generalization abilities. A typical vision-language model architecture consists of a text encoder and an image encoder. During inference, a hand-crafted template prompt is combined with all category inputs to produce text features through the text encoder. These features are then compared with image features, extracted by the image encoder, to calculate similarity and thereby determine the predicted category.

\begin{figure}[t]
 \centering
 \includegraphics[width=0.95\columnwidth]{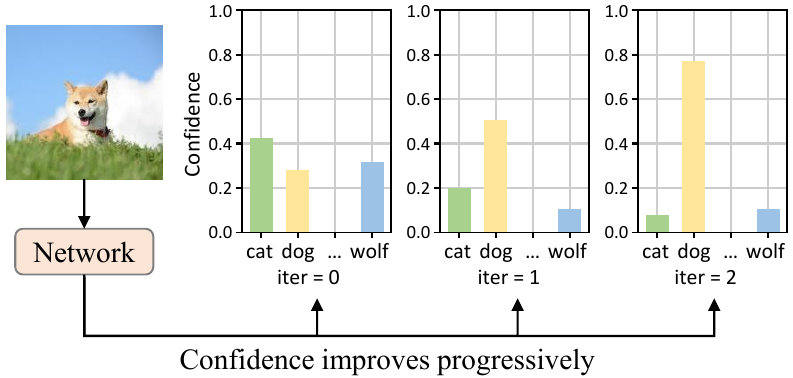}
 \vspace{-0.1in}
 \caption{An illustration of our presented method. ProMPT progressively refines classification confidence, rectifying from ``cat'' to ``dog'' with iterative network processing.}
 \label{fig:fig1}
  \vspace{-0.25in}
\end{figure}
While VLMs exhibit impressive performance in generalizing to new classes, significant challenges arise when fine-tuning them for downstream tasks. Firstly, fine-tuning the whole model demands massive data, since insufficient data results in over-fitting. Secondly, fine-tuning large-scale models requires considerable computational resources and risks the catastrophic forgetting of previously acquired knowledge. Addressing the aforementioned concerns, extensive research \cite{radford2021learning,jin2022good,zhou2022learning,zhou2022conditional} has highlighted prompt learning as an effective paradigm. It originates in the field of natural language processing (NLP) \cite{shin2020autoprompt,petroni2019language,lester2021power,li2021prefix}, encompassing hard prompts and soft prompts. Hard prompts are artificially crafted sentence templates designed to restructure the input to resemble the format in pre-training. However, designing templates requires careful verification with intensive labor as minor wording changes can significantly affect performance \cite{zhou2022learning}. Recently, a series of works \cite{shu2022test,zhu2023prompt,yao2023visual,khattak2023maple} have explored soft prompts for VLMs. Concretely, soft prompts are learnable vectors that are injected into the model to stimulate information beneficial for downstream tasks. A representative method is CoOp \cite{zhou2022learning}, which trains learnable parameters in an end-to-end manner while keeping the entire pre-trained parameters fixed.

Taking into account the studies mentioned, we summarize the following considerations. First, image classification relying on VLMs is a multi-modal task that contends with a significant domain gap between V-L modalities, as well as challenges of data acquisition and annotation. Existing methods typically utilize a one-pass forward pipeline, where text and image encoders separately process input text and images once. The prediction is inferred by the similarity between V-L features. However, owing to the vast gap between image and text, aligning their features effectively is non-trivial. Second, existing prompting methods for VLMs predominantly concentrate on adapting the language branch, leaving the vision branch unchanged. Since the primary objective of VLMs is to harmonize the embeddings in the V-L branches, a single text prompting may hinder modeling of the correlation between the output embeddings in both branches, thus leading to sub-optimal solutions. 

Regarding the issues, we propose \textbf{Pro}gressive \textbf{M}ulti-modal conditional \textbf{P}rompt \textbf{T}uning (ProMPT), a simple yet efficient approach. ProMPT draws inspiration from the work of Feng et al. \cite{feng2021docscanner,feng2024recurrent} and discovers that in the human process of recognizing images, images can be repeatedly and deeply understood. Repetitive digestion of image aids in enhancing classification accuracy. Rather than making a direct prediction, ProMPT revisits the original image multiple times to check answers, incrementally refining the prediction from coarse to precise. In Figure~\ref{fig:fig1}, when an image is input into our network for classification, the confidence of the correct category progressively rises with each iteration. This process enables the prediction to be corrected from initial wrong ``cat'' to eventual correct ``dog''.

Specifically, we implement a recurrent architecture that employs an iterative evolution strategy to align V-L features for accurate prediction. ProMPT comprises two main modules: an initialization module and a multi-modal iterative evolution (MIE) module. In initialization, for a given image, we utilize the original CLIP to encode text and image, generating V-L features. The cosine similarity is then calculated to select the text features of the top-${a}$ categories, which serve as the initial input for MIE. Following that, the V-L features are updated progressively through MIE, containing three steps: class-conditional vision prompting, instance-conditional text prompting, and feature filtering. On one hand, to make vision encoding more focused on the relevant target objects, vision prompts are derived from the top-${a}$ text features through a vision generator. On the other hand, inspired by CoCoOp \cite{zhou2022conditional}, we convert image features into instance-conditional text prompts via a text generator to foster generalization. Feature filtering is intended to select the most image-relevant top-${a}$ text features. Unlike most uni-modal methods, we introduce prompts in both V-L branches to facilitate the alignment of V-L features. Throughout the iterative process, vision and text prompts are continuously optimized, stimulating useful knowledge of VLMs, and promoting better alignment of V-L features. ProMPT evolves results, from rough to accurate prediction.

To evaluate the efficacy of our proposed ProMPT, we conduct comprehensive experiments across three representative settings: generalization from base-to-novel classes, cross-dataset evaluation, and domain generalization. The experimental results demonstrate the superiority of ProMPT over established baselines. Notably, in the generalization from base-to-novel classes setting, our method surpasses the baseline CoCoOp \cite{zhou2022conditional} in 10 out of 11 datasets, achieving absolute performance improvements of 3.2\% in new classes and 1.97\% in harmonic mean (HM). Additionally, in the cross-dataset evaluation and domain generalization settings, ProMPT exhibits valid generalizability and robustness with optimal average accuracy.

\vspace{-0.05in}
\section{RELATED WORKS}
In this section, we provide an overview of the related works, focusing on vision-language models and prompt learning.

% \vspace{-0.05in}
% \subsection{Vision-Language Models}
\smallskip
\textbf{Vision-Language Models.}$\;$
Recently, the field of CV has witnessed the advent and growing application of VLMs, like CLIP \cite{radford2021learning}, ALIGN \cite{jia2021scaling}, and Florence \cite{yuan2021florence}, particularly in few-shot or zero-shot learning scenarios. Vision-language models are trained on corpora of enormous noisy image-text pairs from the web based on contrastive learning by pulling the representations of matching image-text pairs close while pushing those of mismatching pairs far away to learn aligned vision-language representations. Under the supervision of natural language, VLMs exhibit impressive efficacy across a broad spectrum of downstream tasks. However, despite their ability to learn generalized representations, direct application to specific downstream tasks often leads to notable performance drops, posing a substantial challenge. Numerous studies have demonstrated that performance can be enhanced by tailoring VLMs with customized methods for downstream tasks, such as visual recognition \cite{zhou2022learning,zhou2022conditional,gao2023clip,zhang2021tip}, video understanding \cite{ju2022prompting,fang2022transferring,amato2023visione,li2022videoclip,zhuo2022clip4hashing}, and object detection \cite{gu2021open,shi2022proposalclip,du2022learning}. In this work, we present progressive multi-modal conditional prompt tuning for vision-language models to promote image classification tasks under few-shot settings.

\smallskip
\textbf{Prompt Learning.}$\;$
% \vspace{-0.05in}
% \subsection{Prompt Learning}
Prompt learning, originating from the NLP field, is generally classified into hard prompts and soft prompts. Hard prompts \cite{jin2022good,petroni2019language,wallace2019universal,mishra2022cross,le2021many} refer to hand-crafted sentence templates. By inserting input sentences into the templates, pre-trained models mimic the pre-training form in downstream tasks, thereby better eliciting the knowledge learned by the models. In addition, a series of works have employed learnable vectors as pseudo tokens injected into the input or hidden layers of models to participate in the attention computation of Transformer, known as soft prompts \cite{lester2021power,li2021prefix,liu2021p,liu2023gpt,han2022ptr}. These can more effectively extract information useful for downstream tasks from the pre-trained models. 

\begin{figure*}[t]
 \centering
 \includegraphics[width=2.0\columnwidth]{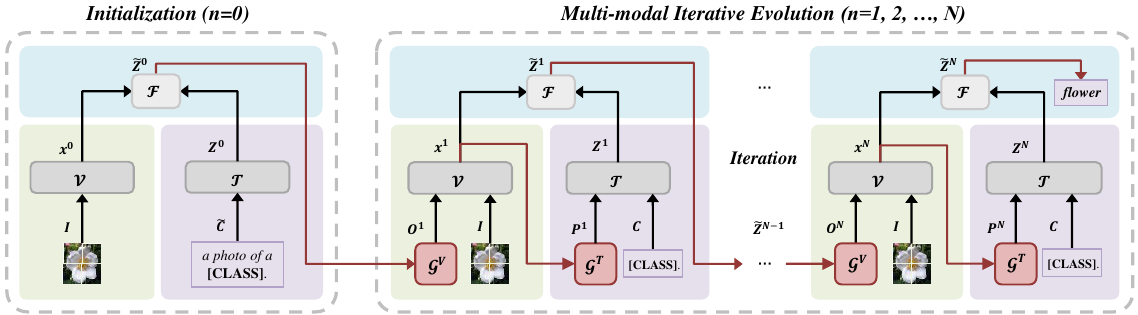}
 \vspace{-0.1in}
 \caption{An overview of our ProMPT framework, adopting an iterative strategy. It comprises an initialization and a multi-modal iterative evolution (MIE) module, aiming to progressively refine the predictions from rough to precise. Initialization contains CLIP and introduces a feature filter to select the top-$a$ text features most similar to image features. Each iteration in MIE involves class-conditional vision prompting, instance-conditional text prompting, and feature filtering. The top-$a$ features are fed into vision generator to produce vision prompts, and then the encoded image features are entered into text generator to obtain text prompts. Overall, ProMPT is optimized by minimizing the cumulative CE loss of the classification in MIE.}
 \label{fig:framework}
  \vspace{-0.1in}
\end{figure*}

In light of the prominent advantages of prompt learning in NLP, numerous approaches \cite{bahng2022exploring,rao2022denseclip,zhu2023prompt,lee2023multimodal,lu2022prompt} have been adopted in the vision and vision-language domains, where the original parameters of the pre-trained model remain unchanged, and only some additional learnable prompt parameters are updated. VPT \cite{jia2022visual} has achieved substantial performance boosts by incorporating a few trainable parameters in the input space while keeping the model backbone frozen. CoOp \cite{zhou2022learning} adapts CLIP-like VLMs for downstream image recognition tasks by modeling word templates as learnable vectors in language branch. CoCoOp \cite{zhou2022conditional} further evolves CoOp \cite{zhou2022learning} by introducing input-conditional prompts for each image, thus enhancing the generalization capabilities of CoOp. TPT \cite{shu2022test} dynamically learns adaptive prompts with a single test sample. DetPro \cite{du2022learning} learns soft prompt representations for open-vocabulary object detection based on VLMs. The above approaches primarily engage with uni-modal prompts either in vision or language branch, restricting VLMs to be optimized in only a single modality and ignoring multi-modal feature interactions. In contrast, we introduce optimizable soft prompts simultaneously in both V-L branches, aiming to advance the alignment between V-L representations.

\section{METHODOLOGY}
In this section, we commence by reviewing the architecture of the pre-trained CLIP. Following this, we briefly outline the overall framework of the proposed method ProMPT. Subsequently, we elaborate on the two main component modules of ProMPT, namely initialization and multi-modal iterative evolution (MIE). Lastly, we detail the design of the training objective tailored for ProMPT.

\subsection{Review of CLIP}
\label{sec:clip}
Our model is constructed upon the foundation of a pre-trained CLIP, which is composed of a text encoder and an image encoder. The text encoder adopts a Transformer \cite{vaswani2017attention} to encode text as vectorized representations, whereas the image encoder proceeds images into feature vectors based on a vision transformer (ViT) \cite{dosovitskiy2020image} or a ResNet \cite{he2016deep}. During the pre-training, Radford et al. \cite{radford2021learning} collect a large number of image-text pairs to implement contrastive learning for CLIP, thus enabling CLIP to learn joint V-L representations. As a result, CLIP is adept at performing zero-shot visual recognition tasks with its parameters entirely frozen. Following existing approaches \cite{zhou2022learning,zhou2022conditional}, our work employs a ViT-based CLIP model. We introduce the encoding process for both vision and text inputs in detail below.

\smallskip
\textbf{Encoding image.}$\;$
The image encoder ${\mathcal{V}}$ comprises $L$ transformer layers ${\{{\mathcal{V}_{l}}\}}_{l=1}^L$ firstly embeds an image $I$ into latent embeddings $ E_0 \in {{\mathbb{R}}^{{M_a} \times d_v }}$. A learnable class token ${s_l} \in {{\mathbb{R}}^{d_v}}$ in image encoder is successively fed into ${\mathcal{V}_{l}}$ along with $ E_{l-1}$ to produce $ E_l$,
\begin{eqnarray}
    [{s_l}, {E_l}] = {\mathcal{V}_{l}}([{s_{l-1}}, {E_{l-1}}]), l=1, 2, ..., L. \label{equ1}
\end{eqnarray}
To derive the final image representation $x$, an image projector layer ${\mathcal{P}_I}$ transforms ${s_L}$ from ${\mathcal{V}_{L}}$ into a shared V-L latent embedding space,
\begin{eqnarray}
    {x} = {\mathcal{P}_I}({s_L}), {x} \in {{\mathbb{R}}^{d}}. \label{equ2}
\end{eqnarray}

\smallskip
\textbf{Encoding text.}$\;$
Given the input text, the text encoder ${\mathcal{T}}$ with $L$ transformer layers ${\{{{\mathcal{T}}_{l}}\}}_{l=1}^L$ tokenizes and embeds it into word embeddings ${{W}_0} = [{w_{0,1}}, {w_{0,2}}, ... , {w_{0,{M_b}}}] \in {{\mathbb{R}}^{({M_b} \times d_l )}}$. At each layer, the embedding ${{W}_{l-1}}$ is input into the $i$-th transformer layer ${\mathcal{T}_l}$ as:
\begin{eqnarray}
    [{{W}_l}] = {{\mathcal{T}}_l}({W}_{l-1}), l=1, 2, ..., L. \label{equ3}
\end{eqnarray}
In a manner analogous to the image branch, the text representation $z$ is derived from the last token ${w_{L,{M_b}}}$ of the last transformer layer through a text projector layer, in the same embedding space as $x$,
\begin{eqnarray}
    {z} = {\mathcal{P}_T}({w_{L,{M_b}}}), {z} \in {{\mathbb{R}}^{d}}. \label{equ4}
\end{eqnarray}

\smallskip
\textbf{Zero-shot inference.}$\;$
During zero-shot inference, the text inputs consist of hand-crafted prompts (e.g., `A photo of a [CLASS]'), where [CLASS] is substituted with the class name of label $y \in [1, . . . , K]$. The score for the $k$-th class is quantified by computing the cosine similarity between the outputs of the text encoder and the image encoder. This calculation employs a similarity function $sim()$ with a temperature parameter $\tau$, expressed as:
\begin{eqnarray}
    {p_k}  = \frac{\exp({sim}(x, {z}_k)/{\tau})}{\sum\nolimits_{i=1}^K\exp({sim}({x},{z}_i)/{\tau})} \label{equ:sim}.
\end{eqnarray}

\subsection{Framework Overview}
To effectively transfer VLMs to image classification tasks, we explore the performance of multi-modal prompting, an advancement over the majority of existing uni-modal prompting approaches. In previous methods \cite{zhou2022learning,zhou2022conditional}, learnable prompts are introduced into the language branch, solely adjusting the text encoding of this branch. However, we hypothesize that limiting prompting to the text encoder alone is sub-optimal. To better align V-L features, we advocate for multi-modal prompt tuning. Drawing inspiration from VPT \cite{jia2022visual}, our method integrates learnable soft prompts into deep layers of ViT in CLIP. Figure~\ref{fig:framework} shows the overall architecture of our proposed ProMPT (\textbf{Pro}gressive \textbf{M}ulti-modal conditional \textbf{P}rompt \textbf{T}uning) framework. In the vision branch, vision prompts are generated by the text features most relevant to the image, encouraging image features to concentrate more on the target objects in the image. Concurrently, in the language branch, we apply image features to generate language prompts. These dynamic prompts have been demonstrated to improve generalizability by Zhou et al \cite{zhou2022conditional}.

Moreover, we mimic the human process of distinguishing images, where an image can be analyzed repeatedly until accurate recognition is achieved. Specifically, we divide ProMPT into two principal phases: initialization ($n=0$) and multi-modal iterative evolution ($n={1, ..., N}$), where $n$ represents the number of iterations. The strategy is designed to incrementally optimize more relevant prompts in each iteration, fostering the alignment of V-L features.

\subsection{Initialization}
For an image classification task with an image $I$ and a set of labels $C = \{{c_k}\}_{k=1}^K$, the initialization phase leverages the original structure of CLIP to encode the input image and text.

ViT ${\mathcal{V}}$ splits $I$ into $M_a$ fixed-size patches which are projected into patch embeddings $ E_0^0 \in {{\mathbb{R}}^{{M_a} \times d_v }}$. Unless otherwise stated, the superscript in the formulas uniformly denotes the iteration number. Subsequently, the process of image encoding in initialization is as follows, similar to Equation~\ref{equ1} and Equation~\ref{equ2},
\begin{eqnarray}
\begin{aligned}
    [{s_l^0}, {E_l^0}] 
    &= {\mathcal{V}_{l}}([{s_{l-1}^0}, {E_{l-1}^0}]), l=1, 2, ..., L,
    \\[1.5mm]
    {x^0} 
    &= {\mathcal{P}_I}({s_L^0}),{x^0} \in {{\mathbb{R}}^{d}}.
\end{aligned}
\end{eqnarray}

In the language branch, the set $C$ is filled into the template prompt to generate $\widetilde{C} = \{\text{a photo of a} \;{c_k}\}_{k=1}^K$. Each category $\widetilde{c}_k$ is tokenized and embedded into word embeddings $ {\widetilde{W}_0^0} = [{P_0}, {{W}_0^0}] \in {{\mathbb{R}}^{({M_b}) \times d_l}}$, where ${P_0} \in {{\mathbb{R}}^{b \times d_l}}$ is the embeddings of the hand-crafted template ``a photo of a'' and ${{W}_0^0} = [{w_{0,1}^0}, ... , {w_{0,({M_b}-{b})}^0}] \in {{\mathbb{R}}^{({M_b}-{b}) \times d_l}}$ represents the embeddings of the category $c_k$. Subsequent to this step, $\widetilde{W}_0^0$ is encoded through each layer of $\mathcal{T}$ to obtain $\widetilde{W}_L^0$ at the last layer $\mathcal{T}_L$. The feature ${w_{L,({M_b}-{b})}^0}$, at the last position of $\widetilde{W}_L^0$, is mapped into text representation ${z_k^0}$ via the text projector layer ${\mathcal{P}_T}$.

\begin{eqnarray}
\begin{aligned}
    [{\widetilde{W}_l^0}] 
    &= {{\mathcal{T}}_l}(\widetilde{W}_{l-1}^0), l=1, 2, ..., L,
    \\[1.5mm]
    {z_k^0} 
    &= {\mathcal{P}_T}({w_{L,({M_b}-{b})}^0}), {z_k^0} \in {{\mathbb{R}}^{d}}.
\end{aligned}
\end{eqnarray}

\begin{figure}[t]
 \centering
 \includegraphics[width=0.65\columnwidth]{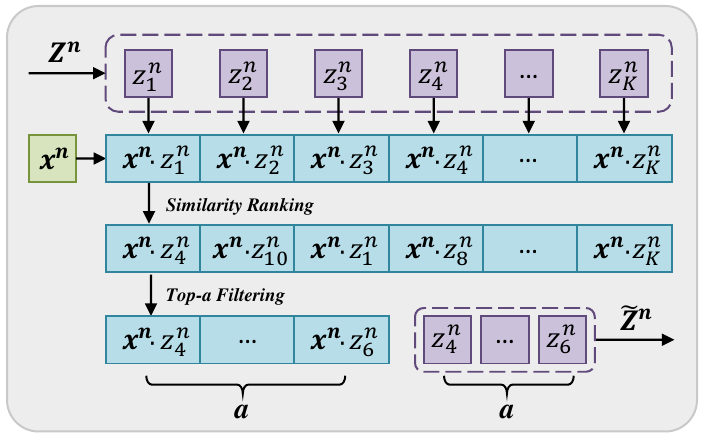}
 \vspace{-0.1in}
 \caption{The implementation process of the feature filter ${\mathcal{F}}$. In the $n$-th iteration, feature filter takes image features ${x^n}$ and text features ${Z^n}$ as input, calculates their cosine similarity, and selects the top-$a$ text features ${\widetilde{Z}}^{n}$ based on the similarity. ${\widetilde{Z}}^{n}$ then serve as the inputs for the $(n+1)$-th iteration.}
 \label{fig:sim}
  \vspace{-0.15in}
\end{figure}

Notably, we incorporate a feature filter $\mathcal{F}$ to extract valuable text information based on $x^0$ and the text representations of the label set ${Z^0} = \{{z_k^0}\}_{k=1}^K$. As depicted in Figure~\ref{fig:sim}, we first calculate the prediction probability $\{ {p_1^0, p_2^0, ..., p_K^0} \}$ for all classes according to Equation~\ref{equ:sim}. Afterwards, the text features ${\widetilde{Z}}^0$ corresponding to the top-${a}$ highest probability values are selected. Later on, ${\widetilde{Z}}^0$ is applied for the generation of conditional prompts during MIE.

\subsection{Multi-modal Iterative Evolution}
CLIP has shown remarkable effectiveness across various tasks, especially in zero-shot scenarios. Therefore, the features of $a$ categories $\widetilde{Z}^0$ we obtained in initialization are approximately valid. We input $\widetilde{Z}^0$ together with image $I$ into the multi-modal iterative evolution (MIE) module, which aims to preserve and enhance the alignment between image and text features. In this way, the V-L features in initialization eventually converge to a precisely aligned state. MIE contains three sub-processes: class-conditional vision prompting, instance-conditional text prompting, and feature filtering, each contributing to the iterative refinement of feature alignment.

\begin{figure}[t]
 \centering
 \includegraphics[width=1.0\columnwidth]{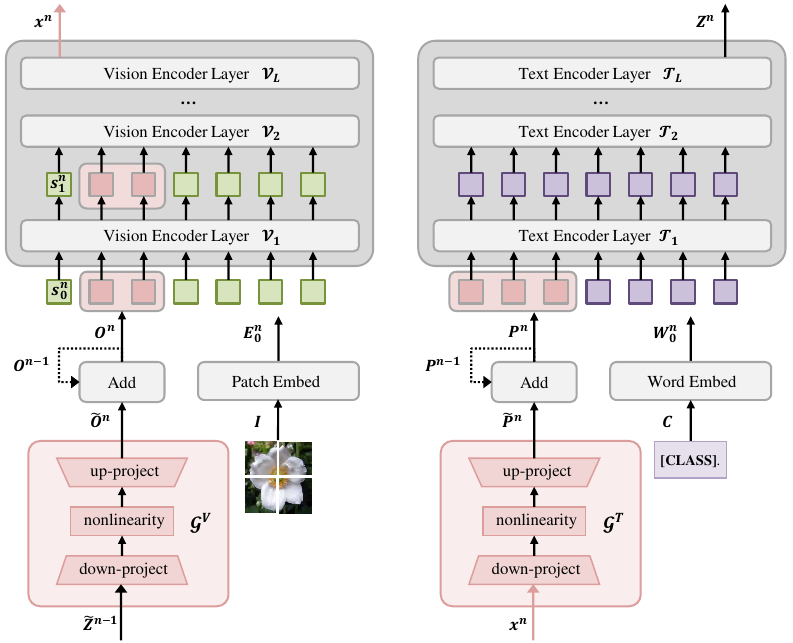}
 \vspace{-0.15in}
 \caption{Class-conditional vision prompting and instance-conditional text prompting at the $n$-th iteration.}
 \label{fig:vgtg}
  \vspace{-0.15in}
\end{figure}

\smallskip
\textbf{Class-conditional vision prompting.}$\;$
To force image features to pay more attention to category-related information during encoding, we implement class-conditional vision prompting. As shown in the left part of Figure~\ref{fig:vgtg}, we introduce prompts ${\{{{O}_j^n} \in {{\mathbb{R}}^{{a} \times {d_{v}}}} \}}_{j=1}^J$ at ${\mathcal{V}_j}$ up to a specific depth $J$ in the $n$-th iteration. Specifically, a set of vision generators ${\mathcal{G}^V} = {\{{\mathcal{G}_{j}^V}\}}_{j=1}^J$ are designed to map $\widetilde{Z}^{n-1}$ into corresponding vision prompts ${{O}_j^n}$ for application in ${\mathcal{V}_j}$. In addition, an Add module is inserted to fully integrate and fuse the class-conditional vision prompts of the $(n-1)$-th iteration process,
\begin{eqnarray}
    {O}_j^n = {{{\mathcal{G}}_j^V}({\widetilde{Z}}^{n-1})} + {O}_j^{n-1},
\end{eqnarray}
where $\mathcal{G}_{j}^V$ is realized through a two-layer MLP (Linear-ReLU-Linear) responsible for mapping text features to image embedding space. The dimensional transformation process is expressed as $d \mapsto \frac{d}{16} \mapsto d_{v}$. Considering that constructing $J$ separate $\mathcal{G}_{j}^V$ would considerably increase the training parameters, we share weight matrix across ${\mathcal{G}^V}$ and set layer-specific bias terms, which is aimed at balancing the training parameters with model performance.

Formally speaking, input embeddings are denoted as $[{s_0^n}, {O_0^n}, {E_0^n}]$. Other vision prompts are further injected in corresponding layers of the image encoder to participate in self-attention computation and the final image representation $x^n$ is computed as follows,
\begin{eqnarray}
\begin{aligned}
    [{s_j^n}, \underline{\hbox to 3.5mm{}}, {E_j^n} \;] 
    & = {\mathcal{V}_{j}}([{s_{j-1}^{n}}, {O_{j}^n}, {E_{j-1}^n}]), j=1, 2, ..., J,
    \\[1.5mm]
    [{s_l^n}, {O_{l+1}^n}, {E_l^n}] 
    & = {\mathcal{V}_{l}}([{s_{l-1}^{n}}, {O_{l}^n}, {E_{l-1}^n}]), l=J+1, 2, ..., L,
    \\[1.5mm]
    {x^n} \
    & = {\mathcal{P}_I}({s_L^n}).
    \end{aligned}
\end{eqnarray}

\smallskip
\textbf{Instance-conditional text prompting.}$\;$
Inspired by CoCoOp \cite{zhou2022conditional}, we apply features $x^n$ from the image branch to generate instance-conditional text prompts via a text generator ${\mathcal{G}^T} = {\{{\mathcal{G}_{r}^T}\}}_{r=1}^R$. As illustrated in the right part of Figure~\ref{fig:vgtg}, the architecture of $\mathcal{G}_r^T$ is the same as that of $\mathcal{G}_j^V$. Instead, we employ a single generator to create text prompts for insertion at the input layer $\mathcal{T}_{1}$, thereby setting $R=1$. The specific process for dimensional transformation follows: ${d} \mapsto \frac{d}{16} \mapsto d_{l}$. With the Add operation, image features $x^n$ pass through ${\mathcal{G}^T}$ to yield text prompts ${{{P}_0^n} \in {{\mathbb{R}}^{{b} \times {d_{l}}}} }$,
\begin{eqnarray}
    {P}_0^n = {{{\mathcal{G}}_1^T}({x}^{n})} + {P}_0^{n-1}.
\end{eqnarray}
Once text prompts are obtained, they are concatenated with ${{W}_0^n}$ and subsequently fed into ${\mathcal{T}_l}$ in turn for computing the text features of the label set ${Z^n} = \{{z_k^n}\}_{k=1}^K$ during the $n$-th iteration.
\begin{eqnarray}
\begin{aligned}
    [{P_{l}^n}, {W_l^n}] 
    & = {\mathcal{T}_{l}}([{P_{l-1}^n}, {W_{l-1}^n}]), l=1, 2, ..., L,
    \\[1.5mm]
    {z_k^n} \
    & = {\mathcal{P}_T}({w_{L,({M_b}-b)}^n}).
\end{aligned}
\end{eqnarray}

\smallskip
\textbf{Feature filtering.}$\;$
In analogy to initialization, the feature filtering of MIE, as depicted in Figure~\ref{fig:sim}, selects the top-$a$ text features ${\widetilde{Z}}^n$ that exhibit the highest relevance to $x^n$ according to prediction probabilities $\{ {p_1^n, p_2^n, ..., p_K^n} \}$ in a similar manner to Equation~\ref{equ:sim}. Subsequently, these filtered features ${\widetilde{Z}}^n$ are utilized as inputs for the next iteration process, continuing the cycle of evolution.
\begin{eqnarray}
    p_k^n = \frac{\exp(\text{sim}(x^n, {z}_k^n)/{\tau})}{\sum\nolimits_{i=1}^K\exp(\text{sim}({x^n},{z}_i^n)/{\tau})} \label{equ:prob} .
\end{eqnarray}

\subsection{Training Objective}
To optimize ProMPT, we adopt a cross-entropy loss function aimed at minimizing the distance between ground-truth labels and prediction probabilities during the $n$-th iteration derived from Equation~\ref{equ:prob},
\begin{eqnarray}
    {\mathcal{L}^n} = -\sum_{k}{{y_k}\,\text{log}\,({p_k^n})}, 1\leq{k}\leq{K},
\end{eqnarray}
where $y_k$ symbolizes the one-hot vector for the ground-truth label. Throughout the training phase, the advanced ProMPT maintains the whole parameters of CLIP fixed, while concurrently engaging in the optimization of prompts and generators. To this end, we apply ${\mathcal{L}^n}$ for the output of all iterations in the MIE, excluding initialization. Overall, the final loss function of our model is formulated as:
\begin{eqnarray}
    {\mathcal{L}} = \sum_{n}{{{\lambda}^{(N-n)}} {\mathcal{L}^n}}, 1\leq{n}\leq{N},
\end{eqnarray}
where $\lambda$ acts as a constant weighting factor for modulating the significance of each iterative evolution. The aggregation of ${\mathcal{L}^n}$ serves to guide the model towards accurate predictions at each iteration, thereby progressively fostering multi-modal learning.

\begin{table*}[htbp]
        \setlength{\tabcolsep}{2.5pt}
	\renewcommand\arraystretch{1.1}
% 	\label{abla}
	\centering
        \normalsize
        \caption{Comparison with state-of-the-art methods in the generalization from base-to-novel classes setting on 11 datasets. ProMPT exhibits vigorous effectiveness by progressively learning multi-modal conditional prompts. H denotes harmonic mean.}
        \vspace{-0.1in}
	\label{tab:result}
	\subtable[\textbf{Average over 11 datasets}]{
            \label{tab:avg}
		\begin{tabular}{lcc|c}
			  \toprule
                       & Base           & New            & H              \\
                \midrule
                CLIP   & 69.34          & 74.22          & 71.70          \\
                CoOp   & \textbf{82.69} & 63.22          & 71.66          \\
                CoCoOp & 80.47          & 71.69          & 75.83          \\
                \midrule
                {ProMPT} & {80.94}      & \textbf{74.89} & \textbf{77.80} \\
                \bottomrule
	\end{tabular}}
	\quad
	\subtable[ImageNet]{
        \label{tab:image}
		\begin{tabular}{lcc|c}
			  \toprule
                       & Base           & New            & H              \\
                \midrule
                CLIP   & 72.43          & 68.14          & 70.22          \\
                CoOp   & \textbf{76.47} & 67.88          & 71.92          \\
                CoCoOp & 75.98          & \textbf{70.43} & \textbf{73.10}          \\
                \midrule
                {ProMPT} & {76.03}      & {70.37} & {73.09} \\
                \bottomrule
	\end{tabular}}
	\quad
	\centering
	\subtable[Caltech101]{
		\begin{tabular}{lcc|c}
			  \toprule
                       & Base           & New            & H              \\
                \midrule
                CLIP   & 96.84          & 94.00          & 95.40          \\
                CoOp   & \textbf{98.00} & 89.81          & 93.73          \\
                CoCoOp & 97.96          & 93.81          & 95.84          \\
                \midrule
                {ProMPT} & {97.77}      & \textbf{94.43} & \textbf{96.07} \\
                \bottomrule
	\end{tabular}}
        \quad
	\subtable[OxfordPets]{
		\begin{tabular}{lcc|c}
			  \toprule
                       & Base           & New            & H              \\
                \midrule
                CLIP   & 91.17          & 97.26          & 94.12         \\
                CoOp   & 93.67          & 95.29          & 94.47          \\
                CoCoOp & 95.20          & 97.69          & 96.43          \\
                \midrule
                {ProMPT} & \textbf{95.20} & \textbf{97.80} & \textbf{96.48} \\
                \bottomrule
	\end{tabular}}
	\vspace{-1pt}
	\subtable[StanfordCars]{
		\begin{tabular}{lcc|c}
			  \toprule
                       & Base           & New            & H              \\
                \midrule
                CLIP   & 63.37          & \textbf{74.89}          & 68.65          \\
                CoOp   & \textbf{78.12}          & 60.40          & 68.13          \\
                CoCoOp & 70.49          & 73.59          & 72.01          \\
                \midrule
                {ProMPT} & 70.57 & 74.00 & \textbf{72.24} \\
                \bottomrule
	\end{tabular}}
	\quad
	\subtable[Flowers102]{
		\begin{tabular}{lcc|c}
			  \toprule
                       & Base           & New            & H              \\
                \midrule
                CLIP   & 72.08          & \textbf{77.80}          & 74.83          \\
                CoOp   & \textbf{97.60} & 59.67          & 74.06          \\
                CoCoOp & 94.87          & 71.75          & 81.71          \\
                \midrule
                {ProMPT} & {94.03}      & 73.03 & \textbf{82.21} \\
                \bottomrule
	\end{tabular}}
	\quad
        \subtable[Food101]{
		\begin{tabular}{lcc|c}
			  \toprule
                       & Base           & New            & H              \\
                \midrule
                CLIP   & 90.10          & 91.22          & 90.66          \\
                CoOp   & 88.33          & 82.26          & 85.19          \\
                CoCoOp & 90.70          & 91.29          & 90.99          \\
                \midrule
                {ProMPT} & \textbf{90.93} & \textbf{91.80} & \textbf{91.36} \\
                \bottomrule
	\end{tabular}}
	\quad
        \subtable[FGVCAircraft]{
		\begin{tabular}{lcc|c}
			  \toprule
                       & Base           & New            & H              \\
                \midrule
                CLIP   & 27.19          & \textbf{36.29} & 31.09          \\
                CoOp   & \textbf{40.44}          & 22.30          & 28.75          \\
                CoCoOp & 33.41          & 23.71          & 27.74          \\
                \midrule
                {ProMPT} & 35.33 & {33.83} & \textbf{34.57} \\
                \bottomrule
	\end{tabular}}
	\vspace{-1pt}
        \subtable[SUN397]{
		\begin{tabular}{lcc|c}
			  \toprule
                       & Base           & New            & H              \\
                \midrule
                CLIP   & 69.36          & 75.35          & 72.23          \\
                CoOp   & \textbf{80.60} & 65.89          & 72.51          \\
                CoCoOp & 79.74          & 76.86          & 78.27          \\
                \midrule
                {ProMPT} & {79.00}      & \textbf{78.10} & \textbf{78.55} \\
                \bottomrule
	\end{tabular}}
        \quad
        \subtable[DTD]{
		\begin{tabular}{lcc|c}
			  \toprule
                       & Base           & New            & H              \\
                \midrule
                CLIP   & 53.24          & \textbf{59.90} & 56.37          \\
                CoOp   & \textbf{79.44} & 41.18          & 54.24          \\
                CoCoOp & 77.01          & 56.00          & 64.85          \\
                \midrule
                {ProMPT} & {77.20} & {57.57} & \textbf{65.95} \\
                \bottomrule
	\end{tabular}}
	\quad
	\subtable[EuroSAT]{
		\begin{tabular}{lcc|c}
			  \toprule
                       & Base           & New            & H              \\
                \midrule
                CLIP   & 56.48          & 64.05          & 60.03          \\
                CoOp   & 92.19          & 54.74          & 68.69          \\
                CoCoOp & 87.49          & 60.04          & 71.21          \\
                \midrule
                {ProMPT} & \textbf{92.47} & \textbf{76.03} & \textbf{83.45} \\
                \bottomrule
	\end{tabular}}
	\quad
	\centering
	\subtable[UCF101]{
		\begin{tabular}{lcc|c}
			  \toprule
                       & Base           & New            & H              \\
                \midrule
                CLIP   & 70.53          & \textbf{77.50} & 73.85          \\
                CoOp   & \textbf{84.69} & 56.05          & 67.46          \\
                CoCoOp & 82.33          & 73.45          & 77.64          \\
                \midrule
                {ProMPT} & {81.80} & {76.80} & \textbf{79.22} \\
                \bottomrule
	\end{tabular}}
\end{table*}

\begin{figure*}[htbp]
\vspace{-0.05in}
 \centering
 \includegraphics[width=1.9\columnwidth]{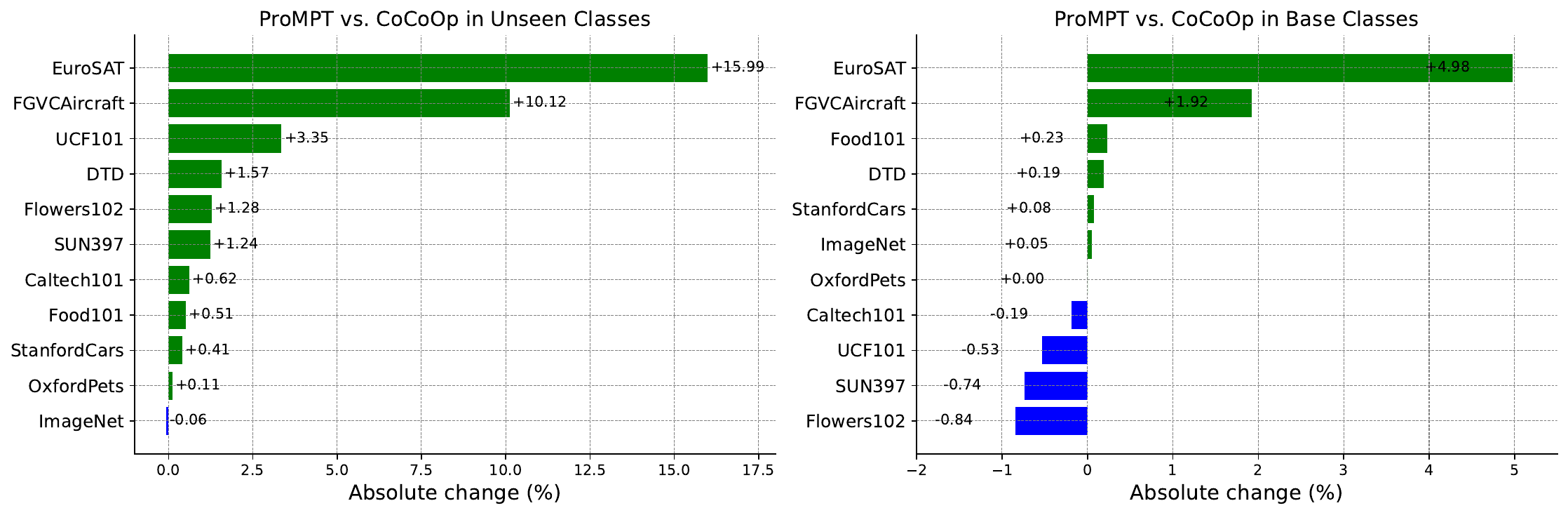}
 \vspace{-0.1in}
 \caption{Comparison of ProMPT and CoCoOp on new and base classes. Generally speaking, ProMPT outperforms CoCoOp on most benchmarks. Some marginal declines are negligible when compared with the considerable enhancements of ProMPT.}
 \label{fig:compare}
  \vspace{-0.05in}
\end{figure*}

\section{EXPERIMENTS}

\subsection{Experimental Setup}
In this section, we elaborate on the experimental setup: benchmark setting, datasets, implementation details, and compared methods.

\smallskip
\textbf{Benchmark setting.}$\;$
Our approach is primarily evaluated in three different experimental settings:
\begin{itemize}
\item[$\bullet$] \textbf{Generalization from base-to-novel classes.}$\;$ To demonstrate the generalizability of ProMPT, we partition datasets into base and novel classes to assess method in a zero-shot setting. The model is first trained on base classes in a few-shot setting and then evaluated on base and novel classes.
\item[$\bullet$] \textbf{Cross-dataset evaluation.}$\;$ To verify the ability of ProMPT in cross-dataset transfer, we train our model on ImageNet in a few-shot manner and directly test it on other datasets.
\item[$\bullet$] \textbf{Domain generalization.}$\;$ To validate the robustness of method on out-of-distribution datasets. Similarly, we deploy our ImageNet-trained model directly on four distinct ImageNet datasets that contain various types of domain shifts.
\end{itemize}

\smallskip
\textbf{Datasets.}$\;$
In our study on generalization from base-to-novel classes setting and cross-dataset evaluation setting, we adopt 11 datasets following Zhou et al. \cite{zhou2022learning}. These include ImageNet \cite{deng2009imagenet}, Caltech101 \cite{fei2004learning}, OxfordPets \cite{parkhi2012cats}, StanfordCars \cite{krause20133d}, Flowers102 \cite{nilsback2008automated}, Food101 \cite{bossard2014food}, FGVCAircraft \cite{maji2013fine}, SUN397 \cite{xiao2010sun}, UCF101 \cite{soomro2012ucf101}, DTD \cite{cimpoi2014describing} and EuroSAT \cite{helber2019eurosat}. These datasets cover a wide range of image classification tasks involving generic objects, fine-grained categories, scene understanding, action recognition, texture classification, and satellite imagery recognition. For assessing domain generalization, we utilize ImageNet as source dataset and its four variants as target datasets, namely ImageNetV2 \cite{recht2019imagenet}, ImageNet-Sketch \cite{wang2019learning}, ImageNet-A \cite{hendrycks2021natural} and ImageNet-R \cite{hendrycks2021many}.

\smallskip
\textbf{Implementation details.}$\;$
Our implementation leverages the open-source CLIP with the ViT-B/16 architecture, where ${d_v} = 768$, ${d_l} = 512$ and $d = 512$. The text prompt length is fixed to 5, with the prompts initialized with the pre-trained CLIP word embeddings of “a photo of a”. The vision prompt length and depth are configured to 8 and 9, respectively. We utilize hyperparameters ${\lambda} = 1.0$ and set iteration numbers $N$ to 2. Experiments are executed on a single GeForce GTX 3090Ti GPU with a batch size of 4 and a learning rate of 0.008 via SGD optimizer for 5 epochs. To streamline the implementation, we uniformly utilize 16 shots for each class. In base-to-novel generalization, we report the accuracies for base and novel classes, alongside their harmonic mean (HM), averaging the outcomes over 3 runs. For cross-dataset evaluation and domain generalization, the model is trained on ImageNet for 2 epochs at a learning rate of 0.0024, with the vision prompt depth adjusted to 3.

\smallskip
\textbf{Compared methods.}$\;$
We evaluate ProMPT and compare it with several notable prompt learning methods for VLMs, including:
\begin{itemize}
\item[$\bullet$] CLIP \cite{radford2021learning} exploits hand-crafted text prompts (``a photo of a [CLASS]''), achieving excellent zero-shot generalization.
\item[$\bullet$] CoOp \cite{zhou2022learning} replaces the hand-crafted prompts with learnable soft prompts that are optimized by the downstream
datasets.
\item[$\bullet$] CoCoOp \cite{zhou2022conditional} introduces Meta-Net on the basis of CoOp, which combines image features with soft prompts of CoOp to generate instance-conditional prompts.
\end{itemize}

\begin{table*}[htbp]
\setlength{\heavyrulewidth}{1pt} % 更粗的顶部和底部线条
\setlength{\tabcolsep}{3.5pt}
\renewcommand\arraystretch{1.1}
\centering
\normalsize
\caption{Comparison of prompt learning methods in the cross-dataset evaluation setting with 16-shot source samples.}
\vspace{-0.05in}
\label{tab:cross_dataset}
\begin{tabular}{lcccccccccccc}
\toprule
\multicolumn{1}{c}{} & Source         & \multicolumn{11}{c}{Target} \\
\cmidrule(r){2-2}   \cmidrule(r){3-13}
\multicolumn{1}{c}{} & ImageNet       & Caltech101     & O-Pets     & S-Cars   & Flowers102     & Food101        & Aircraft   & SUN397         & DTD            & EuroSAT        & UCF101         & \emph{Average}  \\
\midrule
CoOp                 & \textbf{71.51} & 93.70          & 89.14          & 64.51          & 68.71          & 85.30          & 18.47          & 64.15          & 41.92          & 46.39          & 66.55          & 63.88 \\
CoCoOp               & 71.02          & \textbf{94.43} & 90.14          & 65.32          & \textbf{71.88} & 86.06          & 22.94          & \textbf{67.36} & \textbf{45.73} & 45.37          & \textbf{68.21} & 65.74 \\
\midrule
ProMPT               & 70.47          & 93.73          & \textbf{90.20} & \textbf{65.83} & 71.20          & \textbf{86.30} & \textbf{24.57} & 67.33          & 44.27          & \textbf{51.13} & 67.93          & \textbf{66.25} \\
\bottomrule
\end{tabular}
\end{table*}

\begin{table*}[htbp]
% \vspace{0.05in}
\setlength{\heavyrulewidth}{1pt} % 更粗的顶部和底部线条
\setlength{\tabcolsep}{6pt}
\renewcommand\arraystretch{1.1}
\centering
\normalsize
\caption{Comparison of ProMPT with existing approaches in domain generalization under 16-shots setting. “vp” and “tp” denote the vision prompting and text prompting, respectively.}
\vspace{-0.05in}
\label{tab:cross_domain}
\begin{tabular}{lccccccc}
\toprule
       & \multirow{2}{*}{Prompts} & Source         & \multicolumn{5}{c}{Target}                                                          \\
\cmidrule(r){3-3}   \cmidrule(r){4-8}
       &                          & ImageNet       & ImageNetV2     & ImageNet-Sketch & ImageNet-A     & ImageNet-R     & {\emph{Average}}        \\
\midrule
CLIP   & hand-crafted             & 66.73          & 60.83          & 46.15           & 47.77          & 73.96          & 57.18          \\
CoOp   & tp                       & \textbf{71.51} & \textbf{64.20}          & 47.99           & 49.71          & 75.21          & 59.28          \\
CoCoOp & tp                       & 71.02          & 64.07 & 48.75           & 50.63          & 76.18          & 59.91          \\
\midrule
ProMPT & vp+tp                    & 70.47          & 63.97          & \textbf{48.90}  & \textbf{50.97} & \textbf{77.17} & \textbf{60.25} \\
\bottomrule
\end{tabular}
% \vspace{-0.1in}
\end{table*}

\begin{table*}[htbp]
% \vspace{0.1in}
\setlength{\heavyrulewidth}{1pt} % 更粗的顶部和底部线条
\setlength{\tabcolsep}{4.5pt} % 减小列间距
\renewcommand\arraystretch{1.0} % 减小行高
\centering
\small
\caption{Impact of the iteration number $N$ in the multi-modal iterative evolution stage. We report the average accuracy on 11 datasets and five specific datasets, namely ImageNet, OxfordPets, Food101, FGVCAircraft, and EuroSAT, respectively.}
\vspace{-0.03in}
\label{tab:iter}
\begin{tabular}{l|ccc|ccc|ccc|ccc|ccc|ccc}
\toprule
N & \multicolumn{3}{c}{Average} & \multicolumn{3}{c}{ImageNet}     & \multicolumn{3}{c}{OxfordPets} & \multicolumn{3}{c}{Food101} & \multicolumn{3}{c}{FGVCAircraft} & \multicolumn{3}{c}{EuroSAT} \\
\midrule
  & Base    & New     & H       & Base       & New      & H        & Base     & New      & H        & Base     & New      & H        & Base      & New       & H        & Base     & New      & H        \\
0 & 69.34   & 74.77   & 71.95   & 72.43      & 68.14    & 70.22    & 91.17    & 97.26    & 94.12     & 90.10   & 91.22   & 90.66   & 27.19      & \textbf{36.29}    & 31.09   & 56.48     & 64.05     & 60.03  \\
1 & 79.69   & 74.01   & 76.74   & \textbf{76.17}      & \textbf{70.50}    & \textbf{73.22}    & 94.60    & 97.53    & 96.04    & 90.70   & \textbf{91.83}   & 91.26   & 30.50      & 24.87    & 27.40    & 88.57     & \textbf{78.67}     & 83.32 \\
2 & \textbf{80.94}   & \textbf{74.89}   & \textbf{77.80}   & 76.03      & 70.37    & 73.09    & \textbf{95.20}    & \textbf{97.80}    & \textbf{96.48}    & \textbf{90.93}   & 91.80   & \textbf{91.36}   & \textbf{35.33}      & 33.83    & \textbf{34.57}   & \textbf{92.47}     & 76.03     & \textbf{83.45}  \\
\bottomrule
\end{tabular}
\end{table*}

\subsection{Generalization from Base-to-Novel Classes}
We compare the proposed ProMPT with the aforementioned methods on the 11 datasets, and the experimental results are shown in Table~\ref{tab:result}. In all tables, the scale of the accuracy is in percentage. See Table~\ref{tab:result} (a), ProMPT achieves a performance enhancement of 0.47\% on base classes and a significant 3.20\% on new classes respectively over the superior CoCoOp from an average performance perspective. This improvement demonstrates that ProMPT is equipped with solid generalization while ensuring basic performance. We attribute this phenomenon to the proposed MIE module, which gradually optimizes the prediction results in an iterative manner. During each iteration of the MIE module, class-conditional vision prompting promotes image features to focus more intently on the target objects during encoding, thus achieving a better alignment with text features; secondly, instance-conditional text prompting learns text prompts for each image rather than specific to some certain classes. These dynamic prompts are more robust to class shifts. In comparison to CoOp, although ProMPT is slightly inferior by 1.75\% on base classes, its generalization on new classes surges from 63.22\% to 74.89\%. Besides, HM boosts from 71.66\% to 77.80\%, further evidencing the excellent capabilities of ProMPT.

More specifically, ProMPT surpasses CoCoOp on roughly all datasets when transferred to new classes, especially on EuroSAT and FGVCAircraft with satisfactory improvements of 15.99\% and 10.12\%, separately, as visualized in Figure~\ref{fig:compare}. This makes sense because, for fine-grained tasks (FGVCAircraft) or specific tasks (EuroSAT), multi-modal conditional prompting that adapts the model from vision-language modality jointly is more dominant than uni-modal ones. On the other hand, minor performance degradations of ProMPT are observed on 4 out of 11 datasets in the base classes. Nevertheless, these reductions only remain within a mere 0.9\%, which is considered trivial in comparison to the substantial gains afforded by ProMPT. In conclusion, with the aim of striking a balance between accuracy and generalization, we inspect the harmonic mean across all datasets depicted in Table~\ref{tab:result}. This comprehensive analysis reveals that ProMPT consistently demonstrates outstanding capability.

\vspace{-0.1in}
\subsection{Cross-dataset Evaluation}
We assess the cross-dataset generalization ability of ProMPT by learning multi-modal prompts on all the 1000 ImageNet classes and then transferring it directly to the remaining 10 datasets. As shown in Table~\ref{tab:cross_dataset},  we compare the performance of ProMPT with that of CoOp and CoCoOp. On the source dataset, ImageNet, ProMPT achieves comparable performance. On the majority of target datasets, ProMPT exhibits superior performance, outperforming CoOp in all 10 datasets and surpassing CoCoOp in 5 out of 10 datasets. Overall, ProMPT stands out among the comparative methods with the highest average accuracy of 66.25\%. This indicates the effectiveness of our proposed conditional multi-modal prompting and iterative evolution strategy in enhancing generalizability.

\vspace{-0.1in}
\subsection{Domain Generalization}
To verify the domain generalization of ProMPT, we train our model on the source ImageNet dataset and directly evaluate it on four out-of-distribution datasets. As depicted in Table~\ref{tab:cross_domain}, our method consistently outperforms all other methods on variant data and achieves the best average performance across the target data sets. These results highlight that the multi-modal conditional prompts markedly augment the robustness of our model.

\begin{figure}[htbp]
 \centering
 % 第一个图像
 \begin{minipage}[t]{1\columnwidth}
 \centering
 \includegraphics[width=0.83\columnwidth]{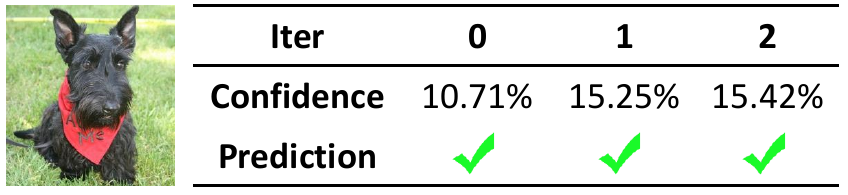}
 \vspace{-0.1in}
 \caption{ProMPT facilitates the refinement of predictions, evolving from initially coarse to precise and accurate classifications. The symbol \ding{52} stands for a right classification.}
 \vspace{0.12in}
 \label{fig:iter1}
 \end{minipage}
 % \hspace{20pt}
 % 第二个图像
 \begin{minipage}[t]{1\columnwidth}
 \centering
 \includegraphics[width=0.83\columnwidth]{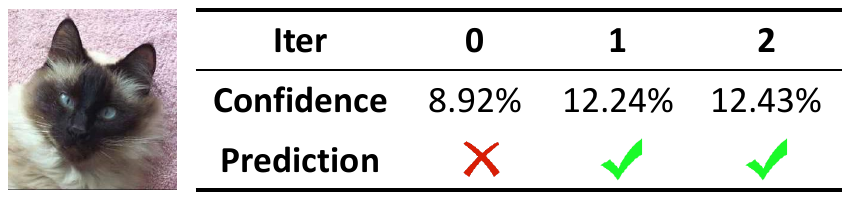}
 \vspace{-0.1in}
 \caption{Through iterative evolution, ProMPT progressively corrects erroneous results made by CLIP to the right ones. \ding{52} and \ding{56} indicate correct and incorrect predictions, respectively.}
 % \vspace{-0.1in}
 \label{fig:iter2}
 \end{minipage}
\end{figure}

\subsection{Ablation Study}
In this section, we put forward several ablation studies to delve deeper into the impact of various factors on ProMPT. We sequentially ablate the number of iterations, model structure, prompt length, vision prompt layer, and loss weighting factor. Unless otherwise specified, the reported results are the average performance of all datasets in the generalization from base-to-novel classes setting. 

\begin{table}[htbp]
\setlength{\heavyrulewidth}{1pt} % 更粗的顶部和底部线条
% \setlength{\lightrulewidth}{1pt} 更粗的中间线条
% \footnotesize % 设置表格字体为 footnotesize
\small
\setlength{\tabcolsep}{7pt} % 减小列间距
\renewcommand{\arraystretch}{0.9} % 减小行高
\caption{Effect of our proposed components in ProMPT. The symbol “-” denotes the exclusion in ProMPT.}
\vspace{-0.1in}
\label{tab:ablation}
\begin{tabular}{l|c|ccccc}
\toprule
Method & ProMPT & -$Iter$ & -${\mathcal{G}^V}$ & -${\mathcal{G}^T}$ & -${\mathcal{G}}$ & -${\mathcal{F}}$    \\
\midrule
Base   & \textbf{80.94} & 69.34 & 80.47 & 79.20 & 78.92 & 78.70 \\
New    & \textbf{74.89} & 74.22 & 73.12 & 70.68 & 70.93 & 72.52 \\
\midrule
H      & \textbf{77.80} & 71.70 & 76.62 & 74.70 & 74.71 & 75.48 \\
\bottomrule
\end{tabular}
\end{table}

\textbf{Effect of iterative optimization.}$\;$
We select several representative datasets, including generic-objects ImageNet and fine-grained OxfordPets, etc., to explore the impact of iteration numbers $N$ on ProMPT by increasing $N$ successively. Table~\ref{tab:iter} lists the results delivered by setting different $N$, suggesting that the average capability improves as iteration continues in training, and reaches a culmination when $N=2$. Furthermore, we visualize the iterative classification process. As illustrated in Figure~\ref{fig:iter1}, it can be observed that with the progression of iterations, ProMPT not only maintains but also gradually raises the confidence for already correctly identified category. Notably, Figure~\ref{fig:iter2} underscores the correction capability of ProMPT. It can rectify category initially misclassified by the basic CLIP to the correct ones, eventually converging to a relatively stable state, thereby achieving stable prediction performance.

\smallskip
\textbf{Effect of main components.}$\;$
We explore the efficacy of each component by removing it from ProMPT. The results are summarized in the Table~\ref{tab:ablation}. Discarding the iterative module (-$Iter$) means only the initialization phase (i.e. CLIP), where the accuracy degrades by 11.6\%, 0.67\%, and 6.1\% on Base, New, and H respectively. Afterwards, we execute three ablation settings: removing vision generator (-${\mathcal{G}^V}$), removing text generator (-${\mathcal{G}^T}$), and removing both (-${\mathcal{G}}$). All settings consistently suffer greater losses in the new classes than the base classes, which suggests that conditional prompts are vital in improving generalization. Finally, we abolish feature filter ${\mathcal{F}}$, which results in considerable damage. Without ${\mathcal{F}}$, all text features are indiscriminately used to generate vision prompts. In this way, image features are equally concerned with some information of mismatched classes during encoding. Hence, ${\mathcal{F}}$ is essential owing to the function of choosing the beneficial information reasonably.

\begin{figure}[t]
 \centering
 \includegraphics[width=1\columnwidth]{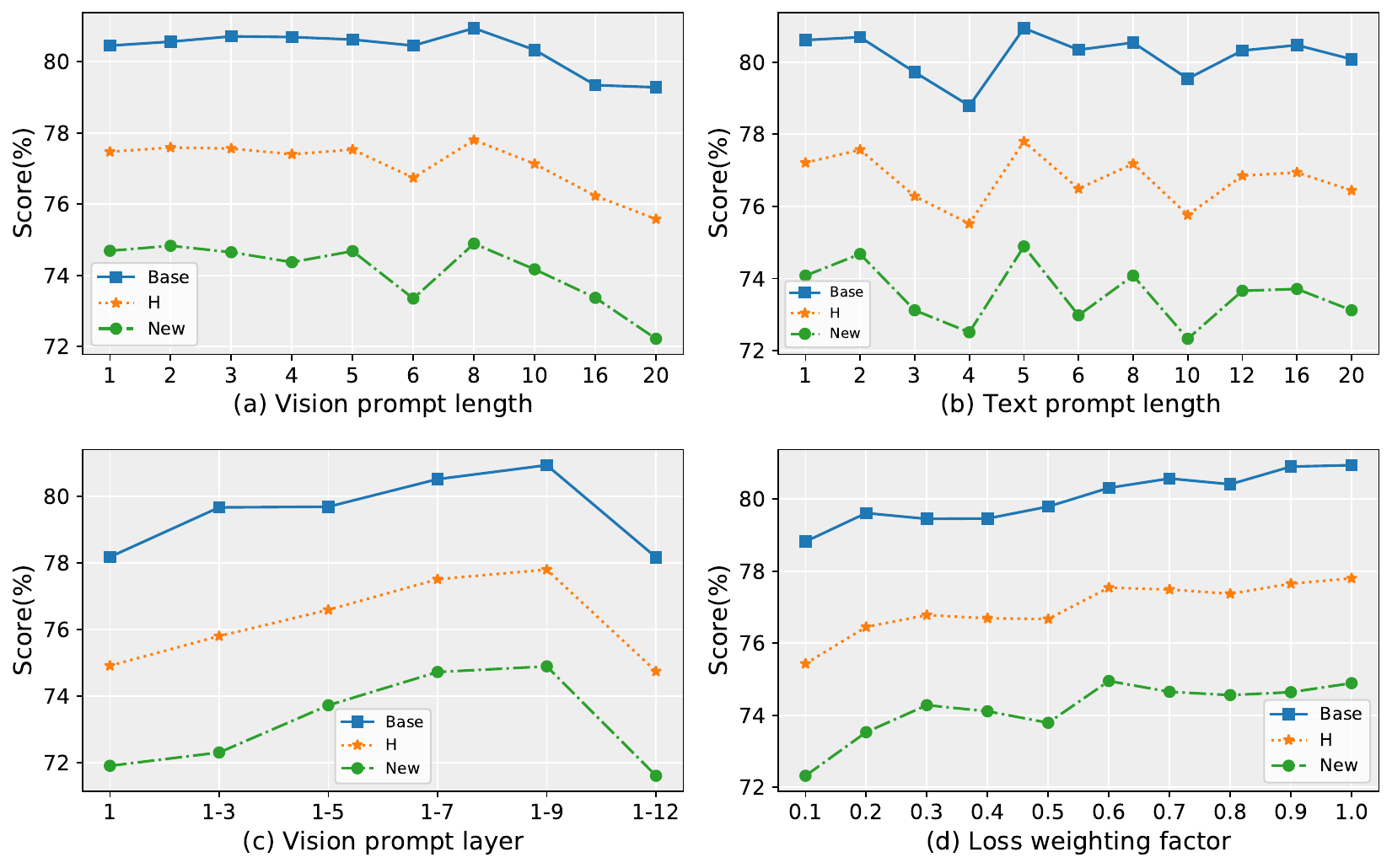}
 \vspace{-0.2in}
 \caption{Ablations of various hyperparameters for ProMPT. Reported results involve accuracy on base and new classes as well as harmonic mean.}
 \label{fig:abla_prompt}
  \vspace{-0.2in}
\end{figure}

\smallskip
\textbf{Effect of prompt length.}$\;$
In Figure~\ref{fig:abla_prompt} (a), we display the effect of vision prompt length $a$ on ProMPT. $a$ determines the number of text features selected in $\mathcal{F}$ and reflects the extent of class-related information included. Additionally, we conduct ablation on text prompt length $b$. Figure~\ref{fig:abla_prompt} (b) presents the results. In brief, the performance initially improves and then decreases with an increase in $b$. Accordingly, we set $a=8$ and $b=5$ for optimal efficiency.

\smallskip
\textbf{Effect of vision prompt layer.}$\;$
We further investigate the impact of varying the number of layers for inserting vision prompts into the vision transformer on ProMPT. It can be intuitively seen from Figure~\ref{fig:abla_prompt} (c) that the more layers of vision prompt are implanted, the better model performance, achieving a peak at 9 layers.

\smallskip
\textbf{Effect of loss weighting factor.}$\;$
We perform ablation for $\lambda$ by varying it. The influence can be inferred from Figure~\ref{fig:abla_prompt} (d) that the effectiveness of ProMPT is positively correlated with $\lambda$. The optimal outcome is observed when ${\lambda} = 1.0$.

\section{CONCLUSION}
In this work, we introduce an innovative framework named Progressive Multi-modal Conditional Prompt Tuning (ProMPT) for VLMs in image classification. To effectively refine and align image and text representations, ProMPT employs a recurrent architecture to leverage the original image and the current encoded information. Further, the presented multi-modal conditional prompt tuning can not only generate class-related vision prompts that make image features concentrate more on the target category, but also yield more robust text prompts suitable for class shifts. In this way, classification results progressively converge from coarse to fine through the multi-modal iterative evolution strategy. Extensive experimental results on three representative settings illustrate the effectiveness of our proposed approach, showcasing better generalization and robustness compared to baselines with a large margin.

%%
%% The acknowledgments section is defined using the "acks" environment
%% (and NOT an unnumbered section). This ensures the proper
%% identification of the section in the article metadata, and the
%% consistent spelling of the heading.
% \begin{acks}
% This work was supported by the GPU cluster built by MCC Lab of Information Science and Technology Institution and the Supercomputing Center of the USTC.
% \end{acks}

%%
%% The next two lines define the bibliography style to be used, and
%% the bibliography file.
\bibliographystyle{ACM-Reference-Format}
\balance
\bibliography{sample-sigconf}

%%% -*-BibTeX-*-
%%% Do NOT edit. File created by BibTeX with style
%%% ACM-Reference-Format-Journals [18-Jan-2012].

\begin{thebibliography}{55}

%%% ====================================================================
%%% NOTE TO THE USER: you can override these defaults by providing
%%% customized versions of any of these macros before the \bibliography
%%% command.  Each of them MUST provide its own final punctuation,
%%% except for \shownote{}, \showDOI{}, and \showURL{}.  The latter two
%%% do not use final punctuation, in order to avoid confusing it with
%%% the Web address.
%%%
%%% To suppress output of a particular field, define its macro to expand
%%% to an empty string, or better, \unskip, like this:
%%%
%%% \newcommand{\showDOI}[1]{\unskip}   % LaTeX syntax
%%%
%%% \def \showDOI #1{\unskip}           % plain TeX syntax
%%%
%%% ====================================================================

\ifx \showCODEN    \undefined \def \showCODEN     #1{\unskip}     \fi
\ifx \showDOI      \undefined \def \showDOI       #1{#1}\fi
\ifx \showISBNx    \undefined \def \showISBNx     #1{\unskip}     \fi
\ifx \showISBNxiii \undefined \def \showISBNxiii  #1{\unskip}     \fi
\ifx \showISSN     \undefined \def \showISSN      #1{\unskip}     \fi
\ifx \showLCCN     \undefined \def \showLCCN      #1{\unskip}     \fi
\ifx \shownote     \undefined \def \shownote      #1{#1}          \fi
\ifx \showarticletitle \undefined \def \showarticletitle #1{#1}   \fi
\ifx \showURL      \undefined \def \showURL       {\relax}        \fi
% The following commands are used for tagged output and should be
% invisible to TeX
\providecommand\bibfield[2]{#2}
\providecommand\bibinfo[2]{#2}
\providecommand\natexlab[1]{#1}
\providecommand\showeprint[2][]{arXiv:#2}

\bibitem[Amato et~al\mbox{.}(2023)]%
        {amato2023visione}
\bibfield{author}{\bibinfo{person}{Giuseppe Amato}, \bibinfo{person}{Paolo Bolettieri}, \bibinfo{person}{Fabio Carrara}, \bibinfo{person}{Fabrizio Falchi}, \bibinfo{person}{Claudio Gennaro}, \bibinfo{person}{Nicola Messina}, \bibinfo{person}{Lucia Vadicamo}, {and} \bibinfo{person}{Claudio Vairo}.} \bibinfo{year}{2023}\natexlab{}.
\newblock \showarticletitle{{VISIONE}: a large-scale video retrieval system with advanced search functionalities}. In \bibinfo{booktitle}{\emph{ICMR}}. \bibinfo{pages}{649--653}.
\newblock


\bibitem[Bahng et~al\mbox{.}(2022)]%
        {bahng2022exploring}
\bibfield{author}{\bibinfo{person}{Hyojin Bahng}, \bibinfo{person}{Ali Jahanian}, \bibinfo{person}{Swami Sankaranarayanan}, {and} \bibinfo{person}{Phillip Isola}.} \bibinfo{year}{2022}\natexlab{}.
\newblock \showarticletitle{Exploring visual prompts for adapting large-scale models}.
\newblock \bibinfo{journal}{\emph{arXiv preprint arXiv:2203.17274}} (\bibinfo{year}{2022}).
\newblock


\bibitem[Bossard et~al\mbox{.}(2014)]%
        {bossard2014food}
\bibfield{author}{\bibinfo{person}{Lukas Bossard}, \bibinfo{person}{Matthieu Guillaumin}, {and} \bibinfo{person}{Luc Van~Gool}.} \bibinfo{year}{2014}\natexlab{}.
\newblock \showarticletitle{Food-101--mining discriminative components with random forests}. In \bibinfo{booktitle}{\emph{ECCV}}. \bibinfo{pages}{446--461}.
\newblock


\bibitem[Cimpoi et~al\mbox{.}(2014)]%
        {cimpoi2014describing}
\bibfield{author}{\bibinfo{person}{Mircea Cimpoi}, \bibinfo{person}{Subhransu Maji}, \bibinfo{person}{Iasonas Kokkinos}, \bibinfo{person}{Sammy Mohamed}, {and} \bibinfo{person}{Andrea Vedaldi}.} \bibinfo{year}{2014}\natexlab{}.
\newblock \showarticletitle{Describing textures in the wild}. In \bibinfo{booktitle}{\emph{CVPR}}. \bibinfo{pages}{3606--3613}.
\newblock


\bibitem[Deng et~al\mbox{.}(2009)]%
        {deng2009imagenet}
\bibfield{author}{\bibinfo{person}{Jia Deng}, \bibinfo{person}{Wei Dong}, \bibinfo{person}{Richard Socher}, \bibinfo{person}{Li-Jia Li}, \bibinfo{person}{Kai Li}, {and} \bibinfo{person}{Li Fei-Fei}.} \bibinfo{year}{2009}\natexlab{}.
\newblock \showarticletitle{{I}mage{N}et: A large-scale hierarchical image database}. In \bibinfo{booktitle}{\emph{CVPR}}. \bibinfo{pages}{248--255}.
\newblock


\bibitem[Dosovitskiy et~al\mbox{.}(2020)]%
        {dosovitskiy2020image}
\bibfield{author}{\bibinfo{person}{Alexey Dosovitskiy}, \bibinfo{person}{Lucas Beyer}, \bibinfo{person}{Alexander Kolesnikov}, \bibinfo{person}{Dirk Weissenborn}, \bibinfo{person}{Xiaohua Zhai}, \bibinfo{person}{Thomas Unterthiner}, \bibinfo{person}{Mostafa Dehghani}, \bibinfo{person}{Matthias Minderer}, \bibinfo{person}{Georg Heigold}, \bibinfo{person}{Sylvain Gelly}, {et~al\mbox{.}}} \bibinfo{year}{2020}\natexlab{}.
\newblock \showarticletitle{An image is worth 16x16 words: Transformers for image recognition at scale}.
\newblock \bibinfo{journal}{\emph{arXiv preprint arXiv:2010.11929}} (\bibinfo{year}{2020}).
\newblock


\bibitem[Du et~al\mbox{.}(2022)]%
        {du2022learning}
\bibfield{author}{\bibinfo{person}{Yu Du}, \bibinfo{person}{Fangyun Wei}, \bibinfo{person}{Zihe Zhang}, \bibinfo{person}{Miaojing Shi}, \bibinfo{person}{Yue Gao}, {and} \bibinfo{person}{Guoqi Li}.} \bibinfo{year}{2022}\natexlab{}.
\newblock \showarticletitle{Learning to prompt for open-vocabulary object detection with vision-language model}. In \bibinfo{booktitle}{\emph{CVPR}}. \bibinfo{pages}{14084--14093}.
\newblock


\bibitem[Fang et~al\mbox{.}(2022)]%
        {fang2022transferring}
\bibfield{author}{\bibinfo{person}{Han Fang}, \bibinfo{person}{Pengfei Xiong}, \bibinfo{person}{Luhui Xu}, {and} \bibinfo{person}{Wenhan Luo}.} \bibinfo{year}{2022}\natexlab{}.
\newblock \showarticletitle{Transferring image-{CLIP} to video-text retrieval via temporal relations}.
\newblock \bibinfo{journal}{\emph{TMM}} (\bibinfo{year}{2022}).
\newblock


\bibitem[Fei-Fei(2004)]%
        {fei2004learning}
\bibfield{author}{\bibinfo{person}{Li Fei-Fei}.} \bibinfo{year}{2004}\natexlab{}.
\newblock \showarticletitle{Learning generative visual models from few training examples: An incremental bayesian approach tested on 101 object categories}. In \bibinfo{booktitle}{\emph{CVPRW}}. IEEE, \bibinfo{pages}{178--178}.
\newblock


\bibitem[Feng et~al\mbox{.}(2024)]%
        {feng2024recurrent}
\bibfield{author}{\bibinfo{person}{Hao Feng}, \bibinfo{person}{Keyi Zhou}, \bibinfo{person}{Wengang Zhou}, \bibinfo{person}{Yufei Yin}, \bibinfo{person}{Jiajun Deng}, \bibinfo{person}{Qi Sun}, {and} \bibinfo{person}{Houqiang Li}.} \bibinfo{year}{2024}\natexlab{}.
\newblock \showarticletitle{Recurrent Generic Contour-based Instance Segmentation with Progressive Learning}.
\newblock \bibinfo{journal}{\emph{TCSVT}} (\bibinfo{year}{2024}).
\newblock


\bibitem[Feng et~al\mbox{.}(2021)]%
        {feng2021docscanner}
\bibfield{author}{\bibinfo{person}{Hao Feng}, \bibinfo{person}{Wengang Zhou}, \bibinfo{person}{Jiajun Deng}, \bibinfo{person}{Qi Tian}, {and} \bibinfo{person}{Houqiang Li}.} \bibinfo{year}{2021}\natexlab{}.
\newblock \showarticletitle{DocScanner: Robust document image rectification with progressive learning}.
\newblock \bibinfo{journal}{\emph{arXiv preprint arXiv:2110.14968}} (\bibinfo{year}{2021}).
\newblock


\bibitem[Gao et~al\mbox{.}(2023)]%
        {gao2023clip}
\bibfield{author}{\bibinfo{person}{Peng Gao}, \bibinfo{person}{Shijie Geng}, \bibinfo{person}{Renrui Zhang}, \bibinfo{person}{Teli Ma}, \bibinfo{person}{Rongyao Fang}, \bibinfo{person}{Yongfeng Zhang}, \bibinfo{person}{Hongsheng Li}, {and} \bibinfo{person}{Yu Qiao}.} \bibinfo{year}{2023}\natexlab{}.
\newblock \showarticletitle{{CLIP}-{A}dapter: Better vision-language models with feature adapters}.
\newblock \bibinfo{journal}{\emph{IJCV}} (\bibinfo{year}{2023}), \bibinfo{pages}{1--15}.
\newblock


\bibitem[Gu et~al\mbox{.}(2021)]%
        {gu2021open}
\bibfield{author}{\bibinfo{person}{Xiuye Gu}, \bibinfo{person}{Tsung-Yi Lin}, \bibinfo{person}{Weicheng Kuo}, {and} \bibinfo{person}{Yin Cui}.} \bibinfo{year}{2021}\natexlab{}.
\newblock \showarticletitle{Open-vocabulary object detection via vision and language knowledge distillation}. In \bibinfo{booktitle}{\emph{ICLR}}.
\newblock


\bibitem[Han et~al\mbox{.}(2022)]%
        {han2022ptr}
\bibfield{author}{\bibinfo{person}{Xu Han}, \bibinfo{person}{Weilin Zhao}, \bibinfo{person}{Ning Ding}, \bibinfo{person}{Zhiyuan Liu}, {and} \bibinfo{person}{Maosong Sun}.} \bibinfo{year}{2022}\natexlab{}.
\newblock \showarticletitle{{PTR}: Prompt tuning with rules for text classification}.
\newblock \bibinfo{journal}{\emph{AI Open}} (\bibinfo{year}{2022}), \bibinfo{pages}{182--192}.
\newblock


\bibitem[He et~al\mbox{.}(2016)]%
        {he2016deep}
\bibfield{author}{\bibinfo{person}{Kaiming He}, \bibinfo{person}{Xiangyu Zhang}, \bibinfo{person}{Shaoqing Ren}, {and} \bibinfo{person}{Jian Sun}.} \bibinfo{year}{2016}\natexlab{}.
\newblock \showarticletitle{Deep residual learning for image recognition}. In \bibinfo{booktitle}{\emph{CVPR}}. \bibinfo{pages}{770--778}.
\newblock


\bibitem[Helber et~al\mbox{.}(2019)]%
        {helber2019eurosat}
\bibfield{author}{\bibinfo{person}{Patrick Helber}, \bibinfo{person}{Benjamin Bischke}, \bibinfo{person}{Andreas Dengel}, {and} \bibinfo{person}{Damian Borth}.} \bibinfo{year}{2019}\natexlab{}.
\newblock \showarticletitle{Euro{SAT}: A novel dataset and deep learning benchmark for land use and land cover classification}.
\newblock \bibinfo{journal}{\emph{JSTARS}} (\bibinfo{year}{2019}), \bibinfo{pages}{2217--2226}.
\newblock


\bibitem[Hendrycks et~al\mbox{.}(2021a)]%
        {hendrycks2021many}
\bibfield{author}{\bibinfo{person}{Dan Hendrycks}, \bibinfo{person}{Steven Basart}, \bibinfo{person}{Norman Mu}, \bibinfo{person}{Saurav Kadavath}, \bibinfo{person}{Frank Wang}, \bibinfo{person}{Evan Dorundo}, \bibinfo{person}{Rahul Desai}, \bibinfo{person}{Tyler Zhu}, \bibinfo{person}{Samyak Parajuli}, \bibinfo{person}{Mike Guo}, {et~al\mbox{.}}} \bibinfo{year}{2021}\natexlab{a}.
\newblock \showarticletitle{The many faces of robustness: A critical analysis of out-of-distribution generalization}. In \bibinfo{booktitle}{\emph{ICCV}}. \bibinfo{pages}{8340--8349}.
\newblock


\bibitem[Hendrycks et~al\mbox{.}(2021b)]%
        {hendrycks2021natural}
\bibfield{author}{\bibinfo{person}{Dan Hendrycks}, \bibinfo{person}{Kevin Zhao}, \bibinfo{person}{Steven Basart}, \bibinfo{person}{Jacob Steinhardt}, {and} \bibinfo{person}{Dawn Song}.} \bibinfo{year}{2021}\natexlab{b}.
\newblock \showarticletitle{Natural adversarial examples}. In \bibinfo{booktitle}{\emph{CVPR}}. \bibinfo{pages}{15262--15271}.
\newblock


\bibitem[Jia et~al\mbox{.}(2021)]%
        {jia2021scaling}
\bibfield{author}{\bibinfo{person}{Chao Jia}, \bibinfo{person}{Yinfei Yang}, \bibinfo{person}{Ye Xia}, \bibinfo{person}{Yi-Ting Chen}, \bibinfo{person}{Zarana Parekh}, \bibinfo{person}{Hieu Pham}, \bibinfo{person}{Quoc Le}, \bibinfo{person}{Yun-Hsuan Sung}, \bibinfo{person}{Zhen Li}, {and} \bibinfo{person}{Tom Duerig}.} \bibinfo{year}{2021}\natexlab{}.
\newblock \showarticletitle{Scaling up visual and vision-language representation learning with noisy text supervision}. In \bibinfo{booktitle}{\emph{ICML}}. \bibinfo{pages}{4904--4916}.
\newblock


\bibitem[Jia et~al\mbox{.}(2022)]%
        {jia2022visual}
\bibfield{author}{\bibinfo{person}{Menglin Jia}, \bibinfo{person}{Luming Tang}, \bibinfo{person}{Bor-Chun Chen}, \bibinfo{person}{Claire Cardie}, \bibinfo{person}{Serge Belongie}, \bibinfo{person}{Bharath Hariharan}, {and} \bibinfo{person}{Ser-Nam Lim}.} \bibinfo{year}{2022}\natexlab{}.
\newblock \showarticletitle{Visual prompt tuning}. In \bibinfo{booktitle}{\emph{ECCV}}. \bibinfo{pages}{709--727}.
\newblock


\bibitem[Jin et~al\mbox{.}(2022)]%
        {jin2022good}
\bibfield{author}{\bibinfo{person}{Woojeong Jin}, \bibinfo{person}{Yu Cheng}, \bibinfo{person}{Yelong Shen}, \bibinfo{person}{Weizhu Chen}, {and} \bibinfo{person}{Xiang Ren}.} \bibinfo{year}{2022}\natexlab{}.
\newblock \showarticletitle{A good prompt is worth millions of parameters: low-resource prompt-based learning for vision-language models}. In \bibinfo{booktitle}{\emph{ACL}}. \bibinfo{pages}{2763--2775}.
\newblock


\bibitem[Ju et~al\mbox{.}(2022)]%
        {ju2022prompting}
\bibfield{author}{\bibinfo{person}{Chen Ju}, \bibinfo{person}{Tengda Han}, \bibinfo{person}{Kunhao Zheng}, \bibinfo{person}{Ya Zhang}, {and} \bibinfo{person}{Weidi Xie}.} \bibinfo{year}{2022}\natexlab{}.
\newblock \showarticletitle{Prompting visual-language models for efficient video understanding}. In \bibinfo{booktitle}{\emph{ECCV}}. \bibinfo{pages}{105--124}.
\newblock


\bibitem[Khattak et~al\mbox{.}(2023)]%
        {khattak2023maple}
\bibfield{author}{\bibinfo{person}{Muhammad~Uzair Khattak}, \bibinfo{person}{Hanoona Rasheed}, \bibinfo{person}{Muhammad Maaz}, \bibinfo{person}{Salman Khan}, {and} \bibinfo{person}{Fahad~Shahbaz Khan}.} \bibinfo{year}{2023}\natexlab{}.
\newblock \showarticletitle{{M}a{PL}e: Multi-modal prompt learning}. In \bibinfo{booktitle}{\emph{CVPR}}. \bibinfo{pages}{19113--19122}.
\newblock


\bibitem[Krause et~al\mbox{.}(2013)]%
        {krause20133d}
\bibfield{author}{\bibinfo{person}{Jonathan Krause}, \bibinfo{person}{Michael Stark}, \bibinfo{person}{Jia Deng}, {and} \bibinfo{person}{Li Fei-Fei}.} \bibinfo{year}{2013}\natexlab{}.
\newblock \showarticletitle{3{D} object representations for fine-grained categorization}. In \bibinfo{booktitle}{\emph{ICCVW}}. \bibinfo{pages}{554--561}.
\newblock


\bibitem[Le~Scao and Rush(2021)]%
        {le2021many}
\bibfield{author}{\bibinfo{person}{Teven Le~Scao} {and} \bibinfo{person}{Alexander~M Rush}.} \bibinfo{year}{2021}\natexlab{}.
\newblock \showarticletitle{How many data points is a prompt worth?}. In \bibinfo{booktitle}{\emph{NAACL}}. \bibinfo{pages}{2627--2636}.
\newblock


\bibitem[Lee et~al\mbox{.}(2023)]%
        {lee2023multimodal}
\bibfield{author}{\bibinfo{person}{Yi-Lun Lee}, \bibinfo{person}{Yi-Hsuan Tsai}, \bibinfo{person}{Wei-Chen Chiu}, {and} \bibinfo{person}{Chen-Yu Lee}.} \bibinfo{year}{2023}\natexlab{}.
\newblock \showarticletitle{Multimodal prompting with missing modalities for visual recognition}. In \bibinfo{booktitle}{\emph{CVPR}}. \bibinfo{pages}{14943--14952}.
\newblock


\bibitem[Lester et~al\mbox{.}(2021)]%
        {lester2021power}
\bibfield{author}{\bibinfo{person}{Brian Lester}, \bibinfo{person}{Rami Al-Rfou}, {and} \bibinfo{person}{Noah Constant}.} \bibinfo{year}{2021}\natexlab{}.
\newblock \showarticletitle{The power of scale for parameter-efficient prompt tuning}. In \bibinfo{booktitle}{\emph{EMNLP}}. \bibinfo{pages}{3045--3059}.
\newblock


\bibitem[Li and Liang(2021)]%
        {li2021prefix}
\bibfield{author}{\bibinfo{person}{Xiang~Lisa Li} {and} \bibinfo{person}{Percy Liang}.} \bibinfo{year}{2021}\natexlab{}.
\newblock \showarticletitle{Prefix-tuning: optimizing continuous prompts for generation}. In \bibinfo{booktitle}{\emph{ACL}}. \bibinfo{pages}{4582--4597}.
\newblock


\bibitem[Li et~al\mbox{.}(2022)]%
        {li2022videoclip}
\bibfield{author}{\bibinfo{person}{Yikang Li}, \bibinfo{person}{Jenhao Hsiao}, {and} \bibinfo{person}{Chiuman Ho}.} \bibinfo{year}{2022}\natexlab{}.
\newblock \showarticletitle{Video{CLIP}: A cross-attention model for fast video-text retrieval task with image clip}. In \bibinfo{booktitle}{\emph{ICMR}}. \bibinfo{pages}{29--33}.
\newblock


\bibitem[Liu et~al\mbox{.}(2021)]%
        {liu2021p}
\bibfield{author}{\bibinfo{person}{Xiao Liu}, \bibinfo{person}{Kaixuan Ji}, \bibinfo{person}{Yicheng Fu}, \bibinfo{person}{Weng~Lam Tam}, \bibinfo{person}{Zhengxiao Du}, \bibinfo{person}{Zhilin Yang}, {and} \bibinfo{person}{Jie Tang}.} \bibinfo{year}{2021}\natexlab{}.
\newblock \showarticletitle{P-tuning v2: Prompt tuning can be comparable to fine-tuning universally across scales and tasks}.
\newblock \bibinfo{journal}{\emph{arXiv preprint arXiv:2110.07602}} (\bibinfo{year}{2021}).
\newblock


\bibitem[Liu et~al\mbox{.}(2023)]%
        {liu2023gpt}
\bibfield{author}{\bibinfo{person}{Xiao Liu}, \bibinfo{person}{Yanan Zheng}, \bibinfo{person}{Zhengxiao Du}, \bibinfo{person}{Ming Ding}, \bibinfo{person}{Yujie Qian}, \bibinfo{person}{Zhilin Yang}, {and} \bibinfo{person}{Jie Tang}.} \bibinfo{year}{2023}\natexlab{}.
\newblock \showarticletitle{{GPT} understands, too}.
\newblock \bibinfo{journal}{\emph{AI Open}} (\bibinfo{year}{2023}).
\newblock


\bibitem[Lu et~al\mbox{.}(2022)]%
        {lu2022prompt}
\bibfield{author}{\bibinfo{person}{Yuning Lu}, \bibinfo{person}{Jianzhuang Liu}, \bibinfo{person}{Yonggang Zhang}, \bibinfo{person}{Yajing Liu}, {and} \bibinfo{person}{Xinmei Tian}.} \bibinfo{year}{2022}\natexlab{}.
\newblock \showarticletitle{Prompt distribution learning}. In \bibinfo{booktitle}{\emph{CVPR}}. \bibinfo{pages}{5206--5215}.
\newblock


\bibitem[Maji et~al\mbox{.}(2013)]%
        {maji2013fine}
\bibfield{author}{\bibinfo{person}{Subhransu Maji}, \bibinfo{person}{Esa Rahtu}, \bibinfo{person}{Juho Kannala}, \bibinfo{person}{Matthew Blaschko}, {and} \bibinfo{person}{Andrea Vedaldi}.} \bibinfo{year}{2013}\natexlab{}.
\newblock \showarticletitle{Fine-grained visual classification of aircraft}.
\newblock \bibinfo{journal}{\emph{arXiv preprint arXiv:1306.5151}} (\bibinfo{year}{2013}).
\newblock


\bibitem[Mishra et~al\mbox{.}(2022)]%
        {mishra2022cross}
\bibfield{author}{\bibinfo{person}{Swaroop Mishra}, \bibinfo{person}{Daniel Khashabi}, \bibinfo{person}{Chitta Baral}, {and} \bibinfo{person}{Hannaneh Hajishirzi}.} \bibinfo{year}{2022}\natexlab{}.
\newblock \showarticletitle{Cross-task generalization via natural language crowdsourcing instructions}. In \bibinfo{booktitle}{\emph{AC}}. \bibinfo{pages}{3470--3487}.
\newblock


\bibitem[Nilsback and Zisserman(2008)]%
        {nilsback2008automated}
\bibfield{author}{\bibinfo{person}{Maria-Elena Nilsback} {and} \bibinfo{person}{Andrew Zisserman}.} \bibinfo{year}{2008}\natexlab{}.
\newblock \showarticletitle{Automated flower classification over a large number of classes}. In \bibinfo{booktitle}{\emph{ICVGIP}}. \bibinfo{pages}{722--729}.
\newblock


\bibitem[Parkhi et~al\mbox{.}(2012)]%
        {parkhi2012cats}
\bibfield{author}{\bibinfo{person}{Omkar~M Parkhi}, \bibinfo{person}{Andrea Vedaldi}, \bibinfo{person}{Andrew Zisserman}, {and} \bibinfo{person}{CV Jawahar}.} \bibinfo{year}{2012}\natexlab{}.
\newblock \showarticletitle{Cats and dogs}. In \bibinfo{booktitle}{\emph{CVPR}}. \bibinfo{pages}{3498--3505}.
\newblock


\bibitem[Petroni et~al\mbox{.}(2019)]%
        {petroni2019language}
\bibfield{author}{\bibinfo{person}{Fabio Petroni}, \bibinfo{person}{Tim Rockt{\"a}schel}, \bibinfo{person}{Sebastian Riedel}, \bibinfo{person}{Patrick Lewis}, \bibinfo{person}{Anton Bakhtin}, \bibinfo{person}{Yuxiang Wu}, {and} \bibinfo{person}{Alexander Miller}.} \bibinfo{year}{2019}\natexlab{}.
\newblock \showarticletitle{Language models as knowledge bases?}. In \bibinfo{booktitle}{\emph{EMNLP-IJCNLP}}. \bibinfo{pages}{2463--2473}.
\newblock


\bibitem[Radford et~al\mbox{.}(2021)]%
        {radford2021learning}
\bibfield{author}{\bibinfo{person}{Alec Radford}, \bibinfo{person}{Jong~Wook Kim}, \bibinfo{person}{Chris Hallacy}, \bibinfo{person}{Aditya Ramesh}, \bibinfo{person}{Gabriel Goh}, \bibinfo{person}{Sandhini Agarwal}, \bibinfo{person}{Girish Sastry}, \bibinfo{person}{Amanda Askell}, \bibinfo{person}{Pamela Mishkin}, \bibinfo{person}{Jack Clark}, {et~al\mbox{.}}} \bibinfo{year}{2021}\natexlab{}.
\newblock \showarticletitle{Learning transferable visual models from natural language supervision}. In \bibinfo{booktitle}{\emph{ICML}}. \bibinfo{pages}{8748--8763}.
\newblock


\bibitem[Rao et~al\mbox{.}(2022)]%
        {rao2022denseclip}
\bibfield{author}{\bibinfo{person}{Yongming Rao}, \bibinfo{person}{Wenliang Zhao}, \bibinfo{person}{Guangyi Chen}, \bibinfo{person}{Yansong Tang}, \bibinfo{person}{Zheng Zhu}, \bibinfo{person}{Guan Huang}, \bibinfo{person}{Jie Zhou}, {and} \bibinfo{person}{Jiwen Lu}.} \bibinfo{year}{2022}\natexlab{}.
\newblock \showarticletitle{{D}ense{CLIP}: Language-guided dense prediction with context-aware prompting}. In \bibinfo{booktitle}{\emph{CVPR}}. \bibinfo{pages}{18082--18091}.
\newblock


\bibitem[Recht et~al\mbox{.}(2019)]%
        {recht2019imagenet}
\bibfield{author}{\bibinfo{person}{Benjamin Recht}, \bibinfo{person}{Rebecca Roelofs}, \bibinfo{person}{Ludwig Schmidt}, {and} \bibinfo{person}{Vaishaal Shankar}.} \bibinfo{year}{2019}\natexlab{}.
\newblock \showarticletitle{Do {I}mage{N}et classifiers generalize to {I}mage{N}et?}. In \bibinfo{booktitle}{\emph{ICML}}. \bibinfo{pages}{5389--5400}.
\newblock


\bibitem[Shi et~al\mbox{.}(2022)]%
        {shi2022proposalclip}
\bibfield{author}{\bibinfo{person}{Hengcan Shi}, \bibinfo{person}{Munawar Hayat}, \bibinfo{person}{Yicheng Wu}, {and} \bibinfo{person}{Jianfei Cai}.} \bibinfo{year}{2022}\natexlab{}.
\newblock \showarticletitle{Proposal{CLIP}: Unsupervised open-category object proposal generation via exploiting {CLIP} cues}. In \bibinfo{booktitle}{\emph{CVPR}}. \bibinfo{pages}{9611--9620}.
\newblock


\bibitem[Shin et~al\mbox{.}(2020)]%
        {shin2020autoprompt}
\bibfield{author}{\bibinfo{person}{Taylor Shin}, \bibinfo{person}{Yasaman Razeghi}, \bibinfo{person}{Robert~L Logan~IV}, \bibinfo{person}{Eric Wallace}, {and} \bibinfo{person}{Sameer Singh}.} \bibinfo{year}{2020}\natexlab{}.
\newblock \showarticletitle{{A}uto{P}rompt: Eliciting knowledge from language models with automatically generated prompts}. In \bibinfo{booktitle}{\emph{EMNLP}}. \bibinfo{pages}{4222--4235}.
\newblock


\bibitem[Shu et~al\mbox{.}(2022)]%
        {shu2022test}
\bibfield{author}{\bibinfo{person}{Manli Shu}, \bibinfo{person}{Weili Nie}, \bibinfo{person}{De-An Huang}, \bibinfo{person}{Zhiding Yu}, \bibinfo{person}{Tom Goldstein}, \bibinfo{person}{Anima Anandkumar}, {and} \bibinfo{person}{Chaowei Xiao}.} \bibinfo{year}{2022}\natexlab{}.
\newblock \showarticletitle{Test-time prompt tuning for zero-Shot generalization in vision-language models}. In \bibinfo{booktitle}{\emph{NeurIPS}}.
\newblock


\bibitem[Soomro et~al\mbox{.}(2012)]%
        {soomro2012ucf101}
\bibfield{author}{\bibinfo{person}{Khurram Soomro}, \bibinfo{person}{Amir~Roshan Zamir}, {and} \bibinfo{person}{Mubarak Shah}.} \bibinfo{year}{2012}\natexlab{}.
\newblock \showarticletitle{UCF101: A dataset of 101 human actions classes from videos in the wild}.
\newblock \bibinfo{journal}{\emph{arXiv preprint arXiv:1212.0402}} (\bibinfo{year}{2012}).
\newblock


\bibitem[Vaswani et~al\mbox{.}(2017)]%
        {vaswani2017attention}
\bibfield{author}{\bibinfo{person}{Ashish Vaswani}, \bibinfo{person}{Noam Shazeer}, \bibinfo{person}{Niki Parmar}, \bibinfo{person}{Jakob Uszkoreit}, \bibinfo{person}{Llion Jones}, \bibinfo{person}{Aidan~N Gomez}, \bibinfo{person}{{\L}ukasz Kaiser}, {and} \bibinfo{person}{Illia Polosukhin}.} \bibinfo{year}{2017}\natexlab{}.
\newblock \showarticletitle{Attention is all you need}. In \bibinfo{booktitle}{\emph{NeurIPS}}. \bibinfo{pages}{6000--6010}.
\newblock


\bibitem[Wallace et~al\mbox{.}(2019)]%
        {wallace2019universal}
\bibfield{author}{\bibinfo{person}{Eric Wallace}, \bibinfo{person}{Shi Feng}, \bibinfo{person}{Nikhil Kandpal}, \bibinfo{person}{Matt Gardner}, {and} \bibinfo{person}{Sameer Singh}.} \bibinfo{year}{2019}\natexlab{}.
\newblock \showarticletitle{Universal adversarial triggers for attacking and analyzing NLP}. In \bibinfo{booktitle}{\emph{EMNLP-IJCNLP}}. \bibinfo{pages}{2153--2162}.
\newblock


\bibitem[Wang et~al\mbox{.}(2019)]%
        {wang2019learning}
\bibfield{author}{\bibinfo{person}{Haohan Wang}, \bibinfo{person}{Songwei Ge}, \bibinfo{person}{Zachary Lipton}, {and} \bibinfo{person}{Eric~P Xing}.} \bibinfo{year}{2019}\natexlab{}.
\newblock \showarticletitle{Learning robust global representations by penalizing local predictive power}.
\newblock \bibinfo{journal}{\emph{NeurIPS}}.
\newblock


\bibitem[Xiao et~al\mbox{.}(2010)]%
        {xiao2010sun}
\bibfield{author}{\bibinfo{person}{Jianxiong Xiao}, \bibinfo{person}{James Hays}, \bibinfo{person}{Krista~A Ehinger}, \bibinfo{person}{Aude Oliva}, {and} \bibinfo{person}{Antonio Torralba}.} \bibinfo{year}{2010}\natexlab{}.
\newblock \showarticletitle{Sun database: Large-scale scene recognition from abbey to zoo}. In \bibinfo{booktitle}{\emph{CVPR}}. \bibinfo{pages}{3485--3492}.
\newblock


\bibitem[Yao et~al\mbox{.}(2023)]%
        {yao2023visual}
\bibfield{author}{\bibinfo{person}{Hantao Yao}, \bibinfo{person}{Rui Zhang}, {and} \bibinfo{person}{Changsheng Xu}.} \bibinfo{year}{2023}\natexlab{}.
\newblock \showarticletitle{Visual-language prompt tuning with knowledge-guided context optimization}. In \bibinfo{booktitle}{\emph{CVPR}}. \bibinfo{pages}{6757--6767}.
\newblock


\bibitem[Yuan et~al\mbox{.}(2021)]%
        {yuan2021florence}
\bibfield{author}{\bibinfo{person}{Lu Yuan}, \bibinfo{person}{Dongdong Chen}, \bibinfo{person}{Yi-Ling Chen}, \bibinfo{person}{Noel Codella}, \bibinfo{person}{Xiyang Dai}, \bibinfo{person}{Jianfeng Gao}, \bibinfo{person}{Houdong Hu}, \bibinfo{person}{Xuedong Huang}, \bibinfo{person}{Boxin Li}, \bibinfo{person}{Chunyuan Li}, {et~al\mbox{.}}} \bibinfo{year}{2021}\natexlab{}.
\newblock \showarticletitle{Florence: A new foundation model for computer vision}.
\newblock \bibinfo{journal}{\emph{arXiv preprint arXiv:2111.11432}} (\bibinfo{year}{2021}).
\newblock


\bibitem[Zhang et~al\mbox{.}(2021)]%
        {zhang2021tip}
\bibfield{author}{\bibinfo{person}{Renrui Zhang}, \bibinfo{person}{Rongyao Fang}, \bibinfo{person}{Wei Zhang}, \bibinfo{person}{Peng Gao}, \bibinfo{person}{Kunchang Li}, \bibinfo{person}{Jifeng Dai}, \bibinfo{person}{Yu Qiao}, {and} \bibinfo{person}{Hongsheng Li}.} \bibinfo{year}{2021}\natexlab{}.
\newblock \showarticletitle{Tip-{A}dapter: Training-free clip-adapter for better vision-language modeling}.
\newblock \bibinfo{journal}{\emph{arXiv preprint arXiv:2111.03930}} (\bibinfo{year}{2021}).
\newblock


\bibitem[Zhou et~al\mbox{.}(2022a)]%
        {zhou2022conditional}
\bibfield{author}{\bibinfo{person}{Kaiyang Zhou}, \bibinfo{person}{Jingkang Yang}, \bibinfo{person}{Chen~Change Loy}, {and} \bibinfo{person}{Ziwei Liu}.} \bibinfo{year}{2022}\natexlab{a}.
\newblock \showarticletitle{Conditional prompt learning for vision-language models}. In \bibinfo{booktitle}{\emph{CVPR}}. \bibinfo{pages}{16816--16825}.
\newblock


\bibitem[Zhou et~al\mbox{.}(2022b)]%
        {zhou2022learning}
\bibfield{author}{\bibinfo{person}{Kaiyang Zhou}, \bibinfo{person}{Jingkang Yang}, \bibinfo{person}{Chen~Change Loy}, {and} \bibinfo{person}{Ziwei Liu}.} \bibinfo{year}{2022}\natexlab{b}.
\newblock \showarticletitle{Learning to prompt for vision-language models}.
\newblock \bibinfo{journal}{\emph{IJCV}} (\bibinfo{year}{2022}), \bibinfo{pages}{2337--2348}.
\newblock


\bibitem[Zhu et~al\mbox{.}(2023)]%
        {zhu2023prompt}
\bibfield{author}{\bibinfo{person}{Beier Zhu}, \bibinfo{person}{Yulei Niu}, \bibinfo{person}{Yucheng Han}, \bibinfo{person}{Yue Wu}, {and} \bibinfo{person}{Hanwang Zhang}.} \bibinfo{year}{2023}\natexlab{}.
\newblock \showarticletitle{Prompt-aligned gradient for prompt tuning}. In \bibinfo{booktitle}{\emph{ICCV}}. \bibinfo{pages}{15659--15669}.
\newblock


\bibitem[Zhuo et~al\mbox{.}(2022)]%
        {zhuo2022clip4hashing}
\bibfield{author}{\bibinfo{person}{Yaoxin Zhuo}, \bibinfo{person}{Yikang Li}, \bibinfo{person}{Jenhao Hsiao}, \bibinfo{person}{Chiuman Ho}, {and} \bibinfo{person}{Baoxin Li}.} \bibinfo{year}{2022}\natexlab{}.
\newblock \showarticletitle{{CLIP4Hashing}: unsupervised deep hashing for cross-modal video-text retrieval}. In \bibinfo{booktitle}{\emph{ICMR}}. \bibinfo{pages}{158--166}.
\newblock


\end{thebibliography}

%%
%% If your work has an appendix, this is the place to put it.
% \appendix

% \section{appendix}

\end{document}